\documentclass[10pt,twocolumn,letterpaper]{article}

\usepackage{cvpr}              

\usepackage{bm}
\usepackage{bbm}
\usepackage{amssymb}
\usepackage{amsthm}
\usepackage{amsmath}
\usepackage{amsfonts}
\usepackage{wasysym}
\usepackage{booktabs}
\usepackage{tabularx}
\usepackage{multirow}
\usepackage{pifont}
\usepackage{verbatim}
\usepackage{makecell}
\usepackage{algorithm}
\usepackage{algpseudocode}
\usepackage{diagbox}
\usepackage[accsupp]{axessibility}
\newtheorem{Definition}{Definition}
\newtheorem{Assumption}{Assumption}
\newtheorem{Theorem}{Theorem}
\newtheorem{Lemma}{Lemma}

\newtheorem{Remark}{Remark}

\usepackage[dvipsnames]{xcolor}

\definecolor{cvprblue}{rgb}{0.21,0.49,0.74}
\usepackage[pagebackref,breaklinks,colorlinks,citecolor=cvprblue]{hyperref}

\def\paperID{9930} 
\def\confName{CVPR}
\def\confYear{2024}

\title{Boosting Adversarial Training via Fisher-Rao Norm-based Regularization}

\author{Xiangyu Yin\quad Wenjie Ruan\thanks{Corresponding Author}\\
University of Liverpool, UK\\
{x.yin22@liverpool.ac.uk,\quad w.ruan@trustai.uk}
}

\begin{document}
\maketitle
\begin{abstract}
Adversarial training is extensively utilized to improve the adversarial robustness of deep neural networks. Yet, mitigating the degradation of standard generalization performance in adversarial-trained models remains an open problem. This paper attempts to resolve this issue through the lens of model complexity. First, We leverage the Fisher-Rao norm, a geometrically invariant metric for model complexity, to establish the non-trivial bounds of the Cross-Entropy Loss-based Rademacher complexity for a ReLU-activated Multi-Layer Perceptron. Then we generalize a complexity-related variable, which is sensitive to the changes in model width and the trade-off factors in adversarial training.
Moreover, intensive empirical evidence validates that this variable highly correlates with the generalization gap of Cross-Entropy loss between adversarial-trained and standard-trained models, especially during the initial and final phases of the training process. Building upon this observation, we propose a novel regularization framework, called Logit-Oriented Adversarial Training (LOAT), which can mitigate the trade-off between robustness and accuracy while imposing only a negligible increase in computational overhead. Our extensive experiments demonstrate that the proposed regularization strategy can boost the performance of the prevalent adversarial training algorithms, including PGD-AT, TRADES, TRADES (LSE), MART, and DM-AT, across various network architectures. Our code will be available at \href{https://github.com/TrustAI/LOAT}{https://github.com/TrustAI/LOAT}.
\end{abstract}    
\section{Introduction}
\label{sec:intro}
Deep Neural Networks (DNNs) are extensively applied in a variety of safety-critical systems,
such as surveillance systems, drones, autonomous vehicles, and malware detection \citep{self-driving,malware,wu2023advdriving,huang2012deep,zhang2023reachability}. Despite their pervasiveness, considerable evidence shows that the standard-trained deep learning models can be easily fooled by subtly modified data points, leading to inaccurate predictions \citep{szegedy2013intriguing,goodfellow2014explaining,mu2021sparse,huang2023survey,yin2022dimba,mu2023certified}. To make models more resistant to such imperceptible perturbations, numerous adversarial training algorithms have been proposed in recent years, such as
PGD-AT \citep{madry2017towards}, TRADES \citep{zhang2019theoretically}, MART \citep{mart}, TRADES (LSE) \citep{Pang2022RobustnessAA}, DM (Diffusion Model)-AT~\cite{wang2023better}, etc. These methods aim to ensure the model's feature representation consistency across clean and adversarial inputs \cite{huang2020survey,wang2022deep}. Nonetheless, it is observed that these adversarial training algorithms commonly lead to a compromise in the standard generalization performance. This phenomenon, often referred to as '{\em trade-off between robustness and accuracy}', continues to be a subject of vigorous discussion in recent years.

Specifically, several studies have attributed the deterioration of standard accuracy
to the bias introduced by adversarial training algorithms \citep{yu2021understanding, Pang2022RobustnessAA}. Other researchers have suggested that sufficient training data is required to bridge the feature gap between adversarial and standard-trained models \citep{tsipras2019robustness,raghunathan2018certified}.
%
Another viewpoint relates this trade-off with the generalizability of the network using a measure known as local Lipschitzness \citep{yang2020closer}. However, most of these studies tend to focus on a {\em single} specific perspective, such as \textit{training objective}, \textit{learning data}, or \textit{model architecture}, which fails to consider the comprehensive integration of these various factors.
This context compels us to explore an important yet challenging question:
\textit{Can we interpret the degradation of standard accuracy in a unified and principled way?}

Affirmatively, as elucidated by \citep{hu2021model}, all the factors above that contribute to the degradation of standard generalization in adversarial training ultimately influence a central concept: {\em model complexity}, which offers a potential approach to analyze this trade-off fundamentally.
Considering the diversity of DNNs, identifying a universal framework for model complexity remains an elusive goal. For the sake of simplicity, we conceptualize models with varying widths and depths as Multi-Layer Perceptrons (MLP). Specifically, given a ReLU-activated $L$-layer MLP, we examine its Rademacher complexity concerning the Cross-Entropy loss out of a set of hypotheses.
The range of complexity is captured by Fisher-Rao norm~\citep{liang2019fisher}, a geometric invariant measure for model complexity. Deriving from this, we can establish bounds for the Rademacher complexity based on the Fisher-Rao norm. 
\begin{figure} 
\centering
\includegraphics[width=\linewidth]{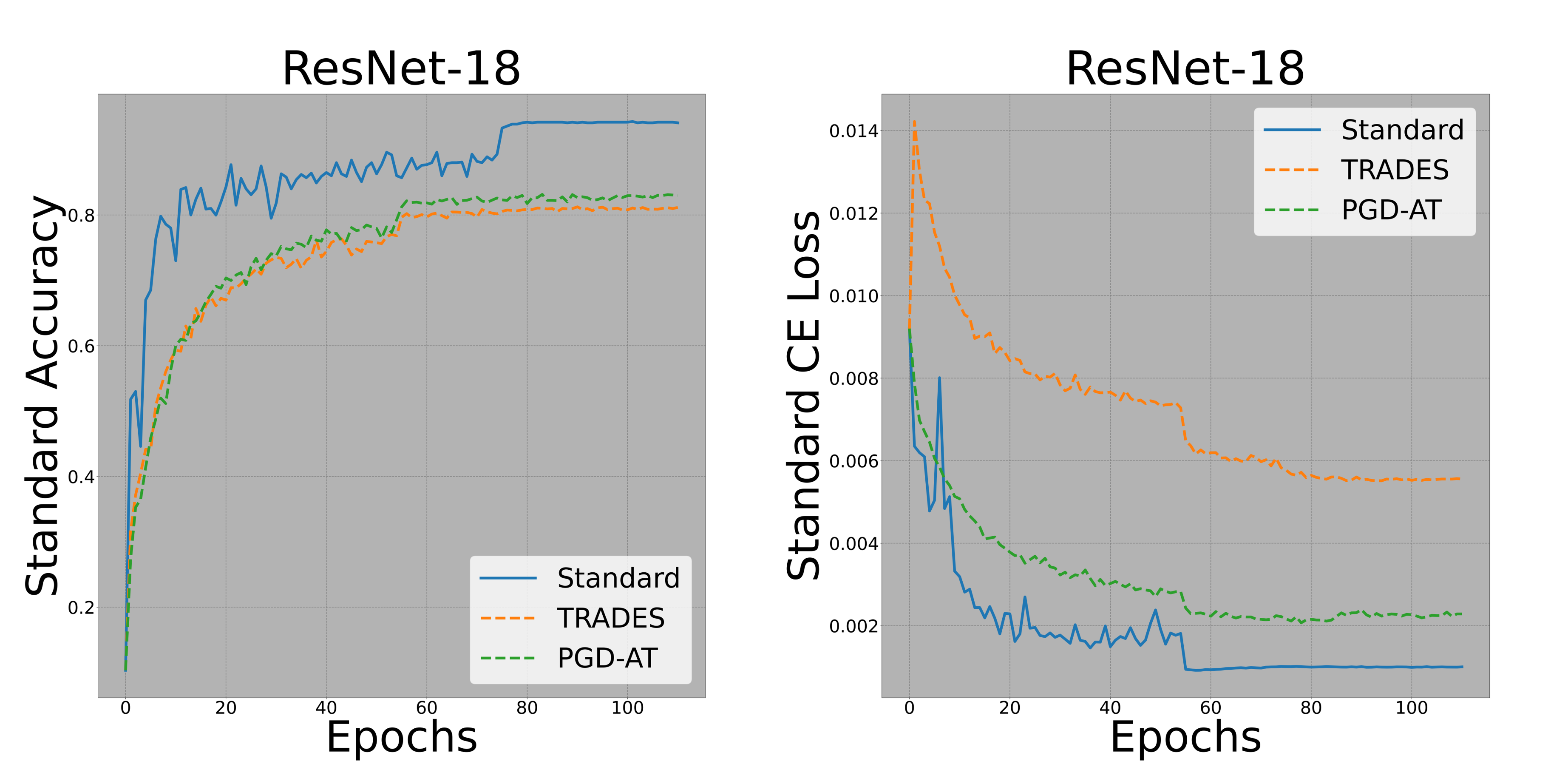}
\caption{Standard generalization performance on CIFAR10.}
\label{figure:motivation}
\end{figure}
Moreover, we observe that the upper and lower bounds of the Cross-Entropy Loss-based Rademacher complexity are significantly influenced by a variable $\Gamma_{ce}$, which is determined by the classification performance on clean training samples. Empirical studies reveal that adversarially-trained MLPs, with different model widths and trade-off factors, have {\em unique} values of $\Gamma_{ce}$. Furthermore, through adjustments to the model width and trade-off factor,
$\Gamma_{ce}$ demonstrates a {\em positive} correlation with the generalization gap of Cross-Entropy loss between adversarial-trained and standard-trained models in the {\em early} epochs, and a {\em negative} correlation in the {\em later} epochs. 

Finally, capitalizing on the empirical link between $\Gamma_{ce}$ and the generalization gap, we propose a novel {\em epoch-specific} framework for adversarial training, named Logit-Oriented Adversarial Training (LOAT). It uniquely combines two regularization tactics: the standard logit-oriented regularizer and the adaptive adversarial logit pairing strategy. Unlike other adversarial training regularizers such as \cite{jin2022enhancing, wu2020adversarial,wang2023self}, LOAT operates as a {\em black-box} solution, preventing the prior knowledge about the model's weights. Notably, LOAT focuses exclusively on the initial and final stages of the adversarial training process, thereby maintaining a {\em minimal} increase in computational demand. We summarize our main contributions as follows:

\begin{itemize}

\item For a ReLU-activated MLP, we introduce the Fisher-Rao norm to capture its Rademacher complexity for Cross-Entropy loss. We empirically and theoretically demonstrate that the logit-based variable $\Gamma_{ce}$ notably influences both the upper and lower bounds.
\item Empirical analysis reveals a non-trivial link between $\Gamma_{ce}$ in an adversarial-trained MLP and critical parameters such as model width and other various trade-off factors. Notably, $\Gamma_{ce}$'s correlation with the generalization gap of Cross-Entropy loss is found to be epoch-dependent, varying across different stages of the training. 
\item We propose a new regularization methodology, Logit-Oriented Adversarial Training (LOAT), which can seamlessly integrate with current adversarial training algorithms and, more importantly, boost their performances without substantially increasing computational overhead.
\end{itemize}

\section{Related works}
\label{sec:related}
\subsection{Trade-off Between Robustness and Accuracy}
Various factors influence the effectiveness of adversarial training \cite{tradeoff-nak,wang2022understanding}. For instance, \citep{tradeoff-nak} highlights the need for increased model capacity to achieve robustness against adversarial examples. 
Other methods utilize data augmentation or mixup to avert robust overfittings  \citep{rice2020overfitting,yin2023rerogcrl}. Additionally, recent studies, including 
\citep{lyu2015unified, ross2017improving, moosavidezfooli2018robustness}, focus on loss perspectives, with strategies like regularization during training to smoothen the input loss landscapes. Conversely, \citep{wu2020adversarial, jin2022enhancing} explore the interplay between weight loss landscape and adversarial robustness.

The above methodologies underscore various means to enhance adversarial robustness. However, as indicated by \citep{tsipras2019robustness, Zhang2019TheoreticallyPT}, there appears to be a natural conflict between adversarial robustness and standard accuracy.  \cite{yu2021understanding} shows that the bias from adversarial training increases with perturbation radius, significantly impacting overall risk. \cite{tsipras2019robustness} observes that the feature representations in adversarial-trained models differ from those in standard-trained models. Alternative perspectives suggest that this trade-off is not inherent and can be mitigated by increasing training data size or theoretically bounding natural and boundary errors. Furthermore, \citep{yang2020closer, Pang2022RobustnessAA} attribute the trade-off to current adversarial training algorithms. Our paper dissects this trade-off, mainly focusing on the influence of model complexity towards the degradation of standard generalization in adversarial training. This phenomenon is depicted in Fig.~\ref{figure:motivation}.

\subsection{Fisher-Rao Norm} 

The study of DNN's capacity has been extensive over the last decade. The Vapnik-Chervonenkis dimension is a classical complexity metric, as discussed in \citep{bartlett2003vapnik}. Yet, its applicability to over-parameterized models might be too broad to explain their generalization. Norm-based measures, as a form of capacity control, have been a focus of recent research \citep{bartlett2017spectrallynormalized, krogh1991simple, neyshabur2015normbased}, although they may not fully encapsulate the distinct variances across various architectures. To address this, the Fisher-Rao norm, introduced in \citep{liang2019fisher}, emerges as a crucial geometric complexity measure. It encompasses existing norm-based capacity metrics and is particularly valued for its geometric invariance, a feature enriched by its connections to information geometry and nonlinear local transformations.

\section{Preliminaries}
\label{sec:preliminaries}
\subsection{Basic Notions}
Consider a typical image classification task, where the objective is to categorize an input image into one of $K$ classes. Let $\mathbf{x}\in[0,1]^d$ represent a normalized input image, and let $y\in\{0,\dots, K-1\}$ denote its corresponding ground truth label. These are assumed to be jointly sampled from a $\mathcal{D}$ distribution. 
Focusing on a specific hypothesis set $\mathcal{F}$ and
a loss function $\mathcal{L}$, the objective of adversarial training is broadly defined as follows:
\begin{equation}
\label{general_objective}
\mathop{\min}_{f\in\mathcal{F}}\left[(1-\lambda)\mathcal{L}(f(\mathbf{x}), y) + \lambda\mathcal{L}(f(\mathcal{A}_{p}^{\epsilon}(\mathbf{x})), y)\right]
\end{equation}
Here, $\mathcal{A}_{p}^{\epsilon}(\mathbf{x})=\mathop{\arg\max}_{\mathbf{x}^{\prime}\in\mathcal{B}_{p}^{\epsilon}(\mathbf{x})}\mathcal{L}(f(\mathbf{x}^{\prime}), y)$, where $\mathcal{B}_{p}^{\epsilon}(\mathbf{x})=\{\mathbf{x}^{\prime}|\left\|\mathbf{x}^{\prime}-\mathbf{x}\right\|_p\leq\epsilon\}$,
$\lambda$ signifies the trade-off factor, and $\epsilon$ represents the radius of the $\ell_p$-norm ball. In particular, when $\lambda$=1.0, it indicates the training objective of PGD-AT. 
Eq.~\ref{general_objective} outlines the process for identifying the optimal hypothesis within 
$\mathcal{F}$. This typically translates to training a neural network that approximates this objective in practical applications.
\begin{Assumption}
For a neural network-based set of hypotheses 
$\mathcal{F}$, every  
$f\in\mathcal{F}$ has an identical architecture and is trained on the same dataset.
\end{Assumption}
Given the recent proliferation of DNN architectures, conducting a theoretical analysis of various hypothesis sets  $\mathcal{F}$ within a universal framework poses a considerable challenge. To address this, our paper simplifies the problem by focusing on MLPs. We analyze the properties of DNNs by varying the depth and width of MLPs, providing a more manageable scope for in-depth study. Detailed definitions are provided below.
\begin{Definition}[$L$-layer MLP]
\label{definition_for_nn}
Given a set of weight matrices $\mathcal{W}=\{\mathbf{W}_1,\cdots,\mathbf{W}_L\}$, and an activation function $\phi(\cdot)$. We define the hypothesis $f(\mathbf{x})$ as an approximation using $L$ layers of matrix multiplication, which is expressed as:
\begin{equation}
f(\mathbf{x})\approx f^{L}_{\mathcal{W}}(\mathbf{x}) = \mathbf{W}_L\phi(\cdots\phi(\mathbf{W}_1\mathbf{x}))
\end{equation}
where 
$\mathbf{W}_{l}\in\mathbb{R}^{H_{l}\times H_{l-1}}$ for $1\leq l\leq L$, and $H_l$ denotes the number of hidden units in the $l$-th layer. Specifically, $H_0 = d$ represents the input dimension, and $H_L$ is the number of output classes.
\end{Definition}
Upon processing by the softmax function $\sigma(\cdot)$, $f_{\mathcal{W}}^L(\mathbf{x})$ yields probabilities for the $K$ classes. Following Eq.~\ref{general_objective} and Def.~\ref{definition_for_nn}, we now present a detailed definition of the risk for clean samples.
\begin{Definition}[Standard Risk]
\label{standard risk}
Consider a specific hypothesis $f^{L}_{\mathcal{W}}$. Empirical risk and population risk for clean samples are defined as follows:
\begin{align}
\label{standard_risk}
\begin{split}
&\tilde{R}_{N_{tr}}\left(\mathcal{L}\circ f_{\mathcal{W}}^{L}\right)=\frac{1}{N_{tr}}\sum_{i=1}^{N_{tr}}\mathcal{L}(f_{\mathcal{W}}^L(\mathbf{x}_i),y_i)\\
&R\left(\mathcal{L}\circ f_{\mathcal{W}}^{L}\right)=\mathbb{E}_{(\mathbf{x},y)\sim\mathcal{D}}\left[\mathcal{L}(f_{\mathcal{W}}^L(\mathbf{x}),y)\right]
\end{split}
\end{align}
$N_{tr}$ denotes the number of clean training samples. 
\end{Definition}
\subsection{Generalization Gap between Algorithms}
Given two distinct training algorithms, $a_1$ and $a_2$, let $\mathcal{F}_{a_1}$ and $\mathcal{F}_{a_2}$ represent the sets of hypotheses generated by each, respectively. Utilizing Def.~\ref{standard risk}, we define the generalization gap of standard risk concerning the loss $\mathcal{L}$ between $\mathcal{F}_{a_1}$ and $\mathcal{F}_{a_2}$ as follows:
\begin{equation}
\small
\label{optimal_gap}
G_{\mathcal{L}}^{\left<\mathcal{F}_{a_1}, \mathcal{F}_{a_2}\right>} = \mathop{\min}_{f_{1}\in\mathcal{F}_{a_1}}R(\mathcal{L}\circ f_1) - \mathop{\min}_{f_2\in\mathcal{F}_{a_2}}R(\mathcal{L}\circ f_2)
\end{equation}
It is important to note that the architecture and training set for $\mathcal{F}_{a_1}$ are identical to those of $\mathcal{F}_{a_2}$, and
$G_{\mathcal{L}}^{\left<\mathcal{F}_{a_1}, \mathcal{F}_{a_2}\right>}$ is determined by training networks in $\mathcal{F}_{a_1}$ and $\mathcal{F}_{a_2}$ for the same number of epochs. For instance, as depicted in Fig.~\ref{figure:motivation}, the generalization gap of Cross-Entropy (CE) loss between adversarial-trained and standard-trained models is observable. This is further characterized by a significant reduction in standard test accuracy across the three adversarial training methods compared to standard training. Specifically, we define the hypothesis sets trained under standard conditions as $\mathcal{F}_{std}$, and those trained using the objective outlined in Eq.~\ref{general_objective} as $\mathcal{F}_{at}$.

\section{Proposed Methods}
\label{sec:methods}
\subsection{Rademacher Complexity via CE Loss}
As depicted in Fig.~\ref{figure:motivation}, $\mathcal{L}_{ce}$ 
  is a key metric for standard test accuracy and is widely used in adversarial training approaches.

In this section, we aim to understand how adversarial training impacts the generalization performance as measured by $\mathcal{L}_{ce}$. To achieve this, we explore the concept of Rademacher Complexity \citep{bartlett2002rademacher}, which affects the ability of the hypothesis set $\mathcal{F}$ to fit random noise and, consequently, its generalization performance.
\begin{Definition} 
\label{rc} Given a neural network-based set of hypotheses $\mathcal{F}$, the Rademacher complexity via $\mathcal{L}$ can be written as:
\small
\begin{equation}
\mathcal{R}_{N_{tr}}\left(\mathcal{L}\circ\mathcal{F}\right) = \mathbb{E}_{\mathbf{\xi}}\frac{1}{N_{tr}}\left[\sup_{f_{\mathcal{W}}^L\in\mathcal{F}}\sum_{i=1}^{N_{tr}}\xi_i\mathcal{L}\left(f_{\mathcal{W}}^L(\mathbf{x}_i), y_i\right)\right]
\end{equation}
\normalsize
where $\xi_i$ takes values in $\{-1, 1\}$ with equal probability. Furthermore, suppose $\mathcal{L}\circ \mathcal{F}\in\left[0, B\right]$, then given $\delta\in(0,1)$, the following generalization bound holds for any $f_{\mathcal{W}}^{L}\in\mathcal{F}$ with probability 1-$\delta$:
\begin{equation}
\label{upper_bound}
\small
R(\mathcal{L}\circ f_{\mathcal{W}}^L)\leq\tilde{R}_{N_{tr}}(\mathcal{L}\circ f_{\mathcal{W}}^L)+2B\mathcal{R}_{N_{tr}}(\mathcal{L}\circ\mathcal{F})+3B\sqrt{\frac{\ln\frac{2}{\delta}}{2N_{tr}}}
\end{equation}
\end{Definition}
\begin{figure}
\centering
\includegraphics[width=\linewidth]{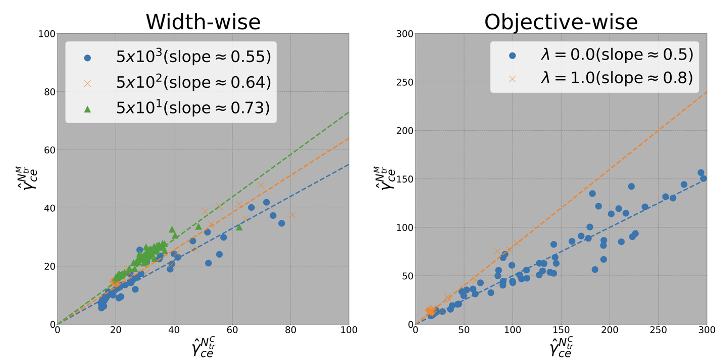}
\caption{The correlation between $\hat{\gamma}_{ce}^{N_{tr}^{C}}$ on the x-axis and $\hat{\gamma}_{ce}^{N_{tr}^{M}}$ on the y-axis within 1-layer MLPs. Each data point corresponds to different epochs during the training process.}
\label{figure:complexity}
\end{figure}
Although it is not feasible to deduce the upper bound for $R(\mathcal{L}_{ce}\circ f_{\mathcal{W}}^{L})$ from Eq.~\ref{upper_bound}, due to the absence of an upper bound for $\mathcal{L}_{ce}$, we can still reference the Rademacher complexity relative to $\mathcal{L}_{ce}$.  It is important to note that a higher Rademacher complexity implies weaker generalization capability, while a lower complexity suggests stronger generalization. Hence, in the ensuing sections, our focus will be on exploring the bounds of $\mathcal{R}_{N_{tr}}(\mathcal{L}_{ce}\circ\mathcal{F})$. 
\subsection{Bounds of complexity via Fisher-Rao Norm}
As detailed in Def.~\ref{rc}, defining the range of $\mathcal{F}$ is crucial for establishing the bounds of Rademacher Complexity in the context of CE loss. Previous research, such as \citep{bartlett2017spectrallynormalized, neyshabur2015normbased, neyshabur2015pathsgd}, has relied on norm-based complexity measures that encompass all possible candidates within $\mathcal{F}$. However, the varied nature of models used in adversarial training, each with its unique architecture, poses a challenge in formulating a standard measure of model complexity. 
\begin{figure}
\centering
\includegraphics[width=\linewidth]{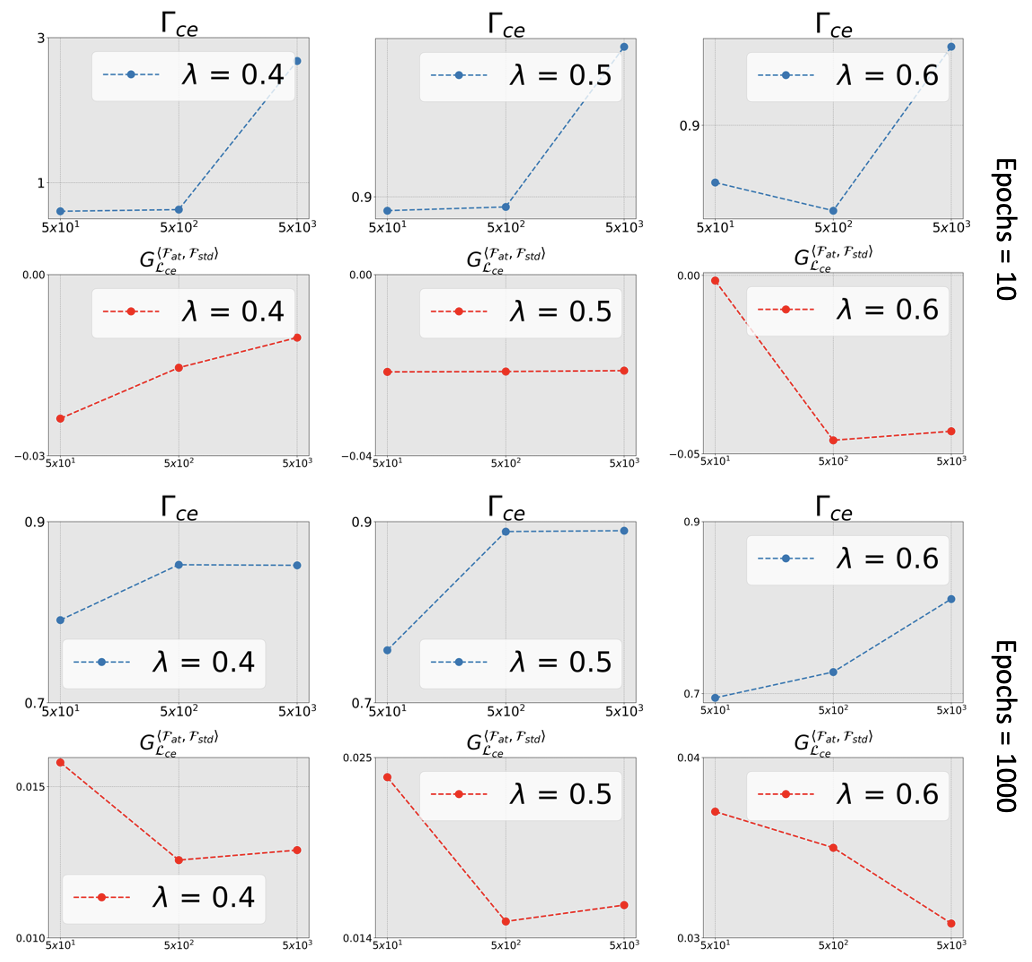}
\caption{Depicting $\Gamma_{ce}$ and $G_{\mathcal{L}_{ce}}^{\left<\mathcal{F}_{at},\mathcal{F}_{std}\right>}$ in 1-layer MLPs with respect to various trade-off factors $\lambda$ ranging from 0.1 to 1.0 in $\mathcal{F}_{at}$. The x-axis represents the number of hidden units from 50 to 5000. }
\label{figure:lambda}
\end{figure}
To address this, we adopt the Fisher-Rao norm, which effectively bypasses the inherent disparities between different models, focusing instead on the variability in output logits.
\begin{Lemma}[Fisher-Rao Norm~\citep{liang2019fisher}]
\label{fisher-rao}
Given an $L$-layer MLP-approximated hypothesis $f_{\mathcal{W}}^{L}$ as defined in Def.~\ref{definition_for_nn}, if $\mathcal{L}$ is smooth with respect to $f_{\mathcal{W}}^{L}$,  the following identity  holds:
\small
\begin{equation}
\label{frn_ori}
\left\|\mathcal{W}\right\|_{FR\circ\mathcal{L}}^2 = L^2\mathbb{E}_{(\mathbf{x},y)\sim\mathcal{D}}\left[\left\langle\frac{\partial\mathcal{L}(f_{\mathcal{W}}^{L}(\mathbf{x}),y)}{\partial f_{\mathcal{W}}^{L}(\mathbf{x})}, f_{\mathcal{W}}^{L}(\mathbf{x})\right\rangle^2\right]
\end{equation}
\end{Lemma}
We then incorporate the CE loss, which exhibits smoothness with respect to $f_{\mathcal{W}}^{L}$, into Lemma~\ref{fisher-rao}. This substitution allows us to establish an upper bound for the radius of the Fisher-Rao norm ball with respect to the CE loss.

\begin{Lemma}[$\mathcal{L}_{ce}$-based Fisher-Rao Norm Ball]
\label{frn}
    Let the radius of the Fisher-Rao norm ball with respect to $\mathcal{L}_{ce}$ be denoted as $\gamma_{ce}=\frac{1}{L}\left\|\mathcal{W}\right\|_{FR\circ\mathcal{L}_{ce}}$. It can be upper bounded by:
    \begin{equation}    \label{frn_ball}\gamma_{ce}\leq\mathbb{E}_{(\mathbf{x},y)\sim\mathcal{D}}\left[\max_{k\neq y}\left|f_{\mathcal{W}}^{L}(\mathbf{x})_k - f_{\mathcal{W}}^{L}(\mathbf{x})_y\right|\right]
    \end{equation}
The right side of the above inequality is denoted as $\hat{\gamma}_{ce}$.
\end{Lemma}
The comprehensive proof of Lemma~\ref{frn} is provided in the Appendix. Both Eq.~\ref{frn_ori} and Eq.~\ref{frn_ball} can be empirically estimated using the training or the test set. Consequently, the Fisher-Rao norm-based complexity measure for the hypothesis $f_{\mathcal{W}}^{L}$ may exhibit variation due to discrepancies between these two datasets. Nevertheless, our analysis is confined to the training set exclusively for the empirical determination of Rademacher complexity bounds. Building on Lemma~\ref{frn}, we proceed to empirically approximate $\hat{\gamma}_{ce}$.
\begin{Definition}\label{radius_frn}
Let $N_{tr}^{C}$ represent the number of correctly classified clean samples and $N_{tr}^{M}$the number of misclassified clean samples. Then, the radii of the Fisher-Rao norm balls concerning
$\mathcal{L}_{ce}$ for $N_{tr}$, $N_{tr}^{C}$, and $N_{tr}^{M}$ can be estimated as follows:
\begin{align}
\label{approximation}
\small
\begin{split}
&\hat{\gamma}_{ce} \approx \frac{1}{N_{tr}}\sum_{i=1}^{N_{tr}}\max_{k\neq y_i}\left|f_{\mathcal{W}}^{L}(\mathbf{x}_i)_{k} - f_{\mathcal{W}}^{L}(\mathbf{x}_i)_{y_i}\right|\\
&\hat{\gamma}_{ce}^{N_{tr}^{C}} \approx \frac{1}{N_{tr}^{C}}\sum_{i=1}^{N_{tr}}\mathcal{I}_{C}\max_{k\neq y_i}\left|f_{\mathcal{W}}^{L}(\mathbf{x}_i)_{k} - f_{\mathcal{W}}^{L}(\mathbf{x}_i)_{y_i}\right|\\
&\hat{\gamma}_{ce}^{N_{tr}^{M}} \approx \frac{1}{N_{tr}^{M}}\sum_{i=1}^{N_{tr}}\mathcal{I}_{M}\max_{k\neq y_i}\left|f_{\mathcal{W}}^{L}(\mathbf{x}_i)_{k} - f_{\mathcal{W}}^{L}(\mathbf{x}_i)_{y_i}\right|
\end{split}
\end{align}
where $\mathcal{I}_C=\mathbbm{1}(\mathop{\arg\max}\sigma(f^L_{\mathcal{W}}(\mathbf{x}_i))=y_i)$, 
$\mathcal{I}_M=\mathbbm{1}(\mathop{\arg\max}\sigma(f^L_{\mathcal{W}}(\mathbf{x}_i))\neq y_i)$
\end{Definition}
In line with Lemma~\ref{frn} and Def.~\ref{radius_frn},  we proceed to examine the Rademacher complexity constrained by the Fisher-Rao norm with respect to $\mathcal{L}_{ce}$.
\begin{Theorem}
Given a set of hypotheses $\mathcal{F}^{\hat{\gamma}_{ce}}=\{f_{\mathcal{W}}^{L}|\left\|\mathcal{W}\right\|_{FR\circ\mathcal{L}_{ce}}\leq L\hat{\gamma}_{ce}\}$, if we denote $\Gamma_{ce} = \frac{\hat{\gamma}_{ce}^{N_{tr}^{C}}-\hat{\gamma}_{ce}^{N_{tr}^M}}{\hat{\gamma}_{ce}^{N_{tr}^{M}}}$, $\mathcal{C}_C=\frac{N_{tr}}{N_{tr}^{C}}$, $\mathcal{C}_M=\frac{N_{tr}}{N_{tr}^{M}}$, $\mathcal{C}_{MC}=\frac{\sqrt{N_{tr}^{M}}+\sqrt{N_{tr}^{C}}}{N_{tr}}$, then we can provide bounds for the Rademacher complexity w.r.t $\mathcal{L}_{ce}$ as follows:
\small
\begin{align}
\label{most_important}
\begin{split}
&\mathcal{R}_{N_{tr}}\left(\mathcal{L}_{ce}\circ\mathcal{F}^{\hat{\gamma}_{ce}}\right)\gtrsim\mathcal{C}_{MC}\ln K-\frac{\hat{\gamma}_{ce}^{N_{tr}^{M}}\mathcal{C}_{C}^{-0.5}(\mathcal{C}_C^{-1}\Gamma_{ce}+1)}{N_{tr}^{0.5}}\\
&\mathcal{R}_{N_{tr}}\left(\mathcal{L}_{ce}\circ\mathcal{F}^{\hat{\gamma}_{ce}}\right)\lesssim\mathcal{C}_{MC}\ln K+\frac{\hat{\gamma}_{ce}^{N_{tr}^{M}}\mathcal{C}_{M}^{-0.5}(\mathcal{C}_C^{-1}\Gamma_{ce}+1)}{N_{tr}^{0.5}}
\end{split}
\end{align}
\normalsize
\label{frn-ce}
\end{Theorem}
\begin{Remark}
\label{remark}
It is noteworthy that the bounds for $\mathcal{R}(\mathcal{L}_{ce}\circ\mathcal{F}^{\hat{\gamma}_{ce}})$ are intricately linked to $\mathcal{C}_{M}^{-0.5}$ and $\mathcal{C}_{C}^{-0.5}$. Theoretically, irrespective of $\mathcal{C}_{MC}\log K
$, the rate of change between the upper and lower bounds with respect to $\Gamma_{ce}$ can be approximated by $\mathcal{O}(\sqrt{\frac{N_{tr}^{M}}{N_{tr}^{C}}})$. Comprehensive proofs of Thm.~\ref{frn-ce} are detailed in the Appendix.
\end{Remark}
\subsection{Sensitivity to Complexity-Related Factors}
\label{gammace_model_complexity}
As noted in Remark~\ref{remark},  Eq.~\ref{most_important} provides the rate of change between the upper and lower bounds of Rademacher complexity with respect to $\Gamma_{ce}$ within a single hypothesis set, yet the variation of $\Gamma_{ce}$ with respect to model complexity-related factors, which result in diverse hypothesis sets preserving apparently different generalization performance, still remains unknown. Therefore, we select two representative model complexity-related factors, model width and training objective, to test the sensitivity of $\Gamma_{ce}$ across different hypothesis sets. Specifically, we train models within various $\mathcal{F}_{at}$ until $\mathcal{L}_{ce}$ reaches convergence.
\begin{figure}
\centering
\includegraphics[width=\linewidth]{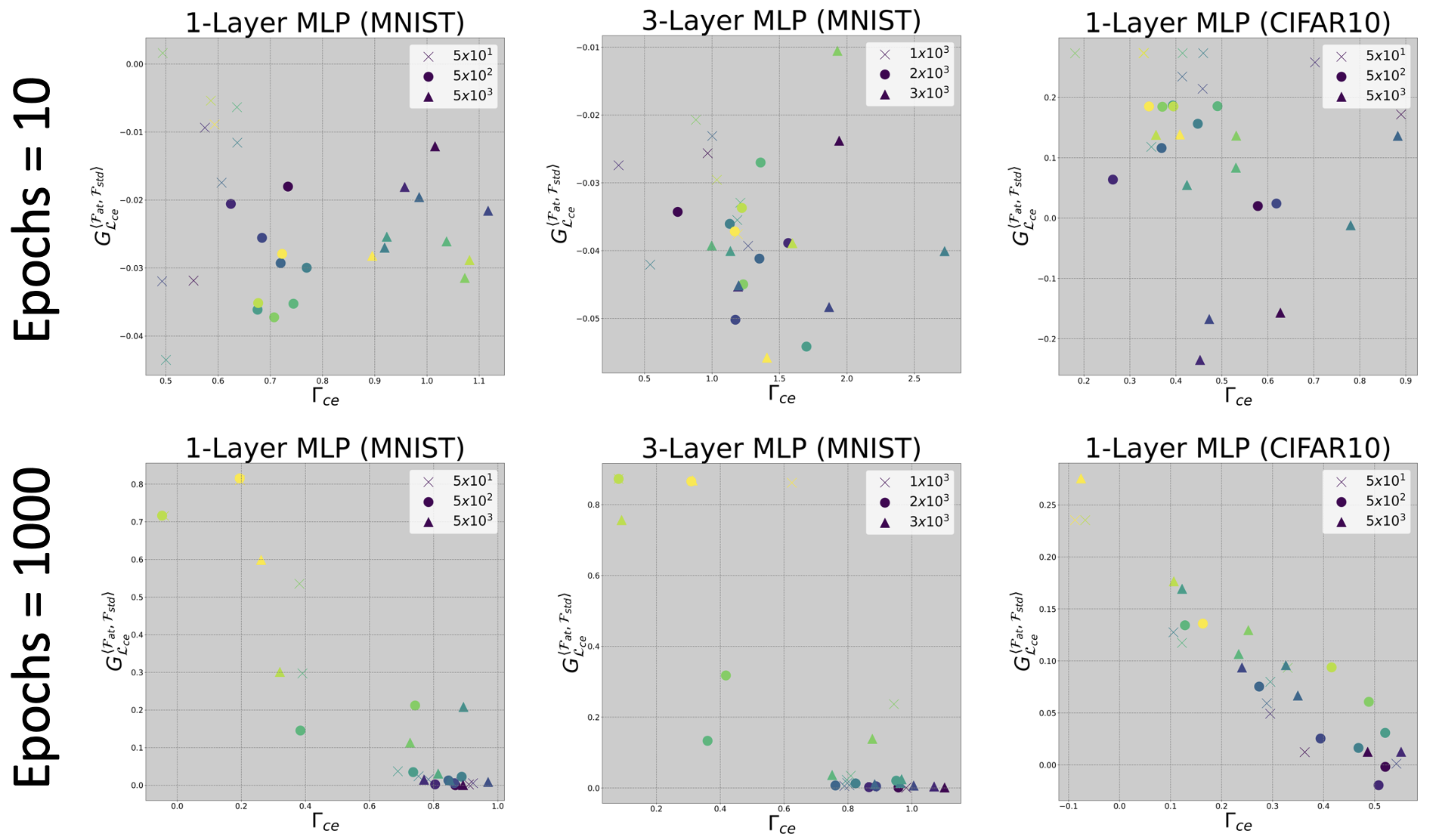}
\caption{Assessment of $G_{\mathcal{L}_{ce}}^{\left<\mathcal{F}_{at}, \mathcal{F}_{std}\right>}$ over three 3 distinct architecture-dataset combinations. Diverse symbols such as $\bm{\times}$, $\blacktriangle$ and $\CIRCLE$ represent different numbers of hidden units.  Additionally, varying shades indicate a range of trade-off factors 
$\lambda$ within $\mathcal{F}_{at}$, specifically from 0.1 to 1.0.}
\label{figure:width-wise}
\end{figure}

\textbf{Width-wise.}
In the left panel of Fig.~\ref{figure:complexity},  we investigate the impact of varying the number of hidden units in a 1-layer MLP, ranging from $5\times 10^{1}$ to $5\times 10^{3}$. An intriguing observation emerges as we increase the number of hidden units: the slope of the dashed line, used to approximate the relationship between $\hat{\gamma}_{ce}^{N_{tr}^{M}}$ and $\hat{\gamma}_{ce}^{N_{tr}^{C}}$, gradually decreases from 0.73 to 0.55. 

\textbf{Objective-wise.} 
In the right panel of Fig.~\ref{figure:complexity}. As  $\lambda$  transitions from 0.0 (indicative of standard training) to 1.0 (corresponding to PGD-AT), the slope of the dashed line shifts from 0.5 to 0.8. 
Specifically, an increase in  $\Gamma_{ce}$, represented by a rising ratio of $\hat{\gamma}_{ce}^{N_{tr}^{C}}/\hat{\gamma}_{ce}^{N_{tr}^{M}}$, is consistent with the trajectory spanning from PGD-AT ($\lambda$=1.0) to standard training ($\lambda$=0.0). 

\subsection{Influence on Generalization Gap of CE Loss}
\label{correlation_empirical}
As shown in Thm.~\ref{frn-ce}, during the early stages of adversarial training on single $\mathcal{F}^{\hat{\gamma}_{ce}}$, characterized by $N_{tr}^{M}\gg N_{tr}^{C}$, an increase in $\Gamma_{ce}$ might lead to a reduction in the lower bound of $\mathcal{R}_{N_{tr}}(\mathcal{L}_{ce}\circ\mathcal{F}^{\hat{\gamma}_{ce}})$. Conversely, a decrease in $\Gamma_{ce}$ tends to have a more pronounced effect on lowering its upper bound. In contrast, when $N_{tr}^{M}\ll N_{tr}^{C}$, although a reduction in $\Gamma_{ce}$ can lower the upper bound of $\mathcal{R}(\mathcal{L}_{ce}\circ\mathcal{F}^{\hat{\gamma}_{ce}})$, an increase in $\Gamma_{ce}$ may cause a more rapid decline in the lower bound. It is commonly known that $G_{\mathcal{L}_{ce}}^{\left<\mathcal{F}_{at}, \mathcal{F}_{std}\right>}$ is significantly influenced by both the model width and the trade-off factor $\lambda$. Therefore, the empirical relationship observed between $\Gamma_{ce}$ and the model width/trade-off factor $\lambda$ as detailed in Sec.~\ref{gammace_model_complexity} leads us to naturally infer that there should also be an inherent correlation between $\Gamma_{ce}$ and $G_{\mathcal{L}_{ce}}^{\left<\mathcal{F}_{at}, \mathcal{F}_{std}\right>}$. 

\textbf{Objective-wise.} Fig.~\ref{figure:lambda} depicts the relationship between $\Gamma_{ce}$ and $G_{\mathcal{L}_{ce}}^{\left<\mathcal{F}_{at}, \mathcal{F}_{std}\right>}$ of 1-layer MLPs in relation to the trade-off factor $\lambda$. For models trained with 10 epochs, $\Gamma_{ce}$ is roughly in positive correlation with $G_{\mathcal{L}_{ce}}^{\left<\mathcal{F}_{at}, \mathcal{F}_{std}\right>}$, whilst for models trained with 1000 epochs,  $\Gamma_{ce}$ is roughly in negative correlation with $G_{\mathcal{L}_{ce}}^{\left<\mathcal{F}_{at}, \mathcal{F}_{std}\right>}$,  which is consistent with the Thm.~\ref{frn-ce}.

\textbf{Width-wise.} Fig.~\ref{figure:width-wise} indicates the relationship between $\Gamma_{ce}$ and  $G_{\mathcal{L}_{ce}}^{\left<\mathcal{F}_{at}, \mathcal{F}_{std}\right>}$ of different MLPs in relation to the number of hidden units. For models trained over 10 epochs, $\Gamma_{ce}$ appears to be broadly in positive correlation with $G_{\mathcal{L}_{ce}}^{\left<\mathcal{F}_{at}, \mathcal{F}_{std}\right>}$, whereas for models subjected to 1000 epochs of training, $\Gamma_{ce}$ tends to exhibit a general negative correlation with $G_{\mathcal{L}_{ce}}^{\left<\mathcal{F}_{at}, \mathcal{F}_{std}\right>}$.

\subsection{Logit-Oriented Adversarial Training}
In Section~\ref{correlation_empirical}, we report a noteworthy trend: $\Gamma_{ce}$ initially correlates positively with the generalization gap of CE loss, but this correlation turns negative as training progresses. To counteract this, we introduce an epoch-dependent regularization strategy. Specifically, we initially penalize the disparity $\hat{\gamma}_{ce}^{N_{tr}^{C}} - \hat{\gamma}_{ce}^{N_{tr}^{M}}$ and later reverse this to penalize $\hat{\gamma}_{ce}^{N_{tr}^{M}} - \hat{\gamma}_{ce}^{N_{tr}^{C}}$, which is in the opposite direction.  Concurrently, we employ an adaptive technique for enhanced model resilience to synchronize the distributions of standard and adversarial logits. These complementary regularization methods construct our proposed Logit-Oriented Adversarial Training (LOAT) framework.
\subsubsection{Standard Logit-Oriented Regularization}
\label{reg_std_logits}
For broader applicability, we represent the adversarial-trained model $f_{\mathcal{W}}^{L}(\cdot)$ by a more general notation, $\mathcal{M}(\cdot)$, and denote the number of training samples, which could be a batch or the entire dataset, as $N$. It is important to note that directly optimizing $\hat{\gamma}_{ce}^{N^M}$ and $\hat{\gamma}_{ce}^{N^C}$ is infeasible due to inherent computational complexities and challenges. Therefore, in line with Eq.~\ref{approximation}, we opt to utilize their lower bounds as the subject to be optimized.
\begin{align}
\scriptsize
\begin{split}
\label{lower_bound}
\overline{\gamma}_{ce}^{N^M}&\approx\frac{1}{N^M}\sum_{i=1}^{N}\mathcal{I}_{M}\left[\frac{1}{K-1}\sum_{k\neq y_i}\left|\mathcal{M}(\mathbf{x}_i)_k - \mathcal{M}(\mathbf{x}_i)_{y_i}\right|\right]\leq\hat{\gamma}_{ce}^{N^M}\\
\overline{\gamma}_{ce}^{N^C}
&\approx\frac{1}{N^C}\sum_{i=1}^{N}\mathcal{I}_{C}\left[\frac{1}{K-1}\sum_{k\neq y_i}\left|\mathcal{M}(\mathbf{x}_i)_k - \mathcal{M}(\mathbf{x}_i)_{y_i}\right|\right]\leq\hat{\gamma}_{ce}^{N^C}\\
\end{split}
\end{align}
where $\mathcal{I}_{C}=\mathbbm{1}(\mathop{\arg\max}\sigma(\mathcal{M}(\mathbf{x}_i))=y_i)$, $\mathcal{I}_{M}=\mathbbm{1}(\mathop{\arg\max}\sigma(\mathcal{M}(\mathbf{x}_i))\neq y_i)$. Echoing the TRADES (LSE) approach ~\cite{Pang2022RobustnessAA}, we use softmax-transformed surrogates instead of raw logits throughout the regularization phase. Moreover, given the non-smooth properties of the $\ell_1$ norm discussed in Eq.~\ref{lower_bound} and its inclination towards generating sparse solutions,  we prefer the $\ell_2$ norm as a more appropriate option. Consequently, following these adjustments, we define our surrogate optimization term $\mathcal{P}_{C}^N$ for $\overline{\gamma}_{ce}^{N^C}$ as follows:
\begin{equation}
\label{approx_correct}
\small
\frac{1}{N^C}\sum_{i=1}^{N}\mathcal{I}_{C}\frac{1}{K-1}\sum_{k\neq y_i}\left(\sigma\left(\mathcal{M}(\mathbf{x}_i)\right)_k - \sigma\left(\mathcal{M}(\mathbf{x}_i)\right)_{y_i}\right)^2
\end{equation}
Notice that for correctly classified samples, it is trivial to prove that $\sigma(\mathcal{M}(\mathbf{x}_i)_{y_i})\geq\frac{1}{K-1}\sum_{k\neq y_i}\sigma(\mathcal{M}(\mathbf{x}_i))_k$, then the lower bound $\check{\mathcal{P}}_C^N$ for $\mathcal{P}_C^N$ can be written as:
\begin{equation}
\label{regularize_correct_bound}
\scriptsize
\frac{1}{N^C}\sum_{i=1}^{N}\mathcal{I}_{C}\frac{1}{K-1}\sum_{k\neq y_i}\left(\sigma\left(\mathcal{M}(\mathbf{x}_i)\right)_k - \frac{1}{K-1}\sum_{k\neq y_i}\sigma\left(\mathcal{M}(\mathbf{x}_i)\right)_{k}\right)^2
\end{equation}
In a parallel manner, we designate the surrogate regularization term for $\overline{\gamma}_{ce}^{N^M}$ as $\mathcal{P}_{M}^N$. In this context, the ensuing inequalities can be established
\begin{align}
\label{regularize_wrong_bound}
\tiny
\begin{split}
&\mathcal{P}_{M}^{N} = \frac{1}{N^M}\sum_{i=1}^{N}\mathcal{I}_{M}\frac{1}{K-1}\sum_{k\neq y_i}\left(\sigma\left(\mathcal{M}(\mathbf{x}_i)\right)_k - \sigma\left(\mathcal{M}(\mathbf{x}_i)\right)_{y_i}\right)^2\\
&\stackrel{(a)}{\geq}\frac{1}{N^M}\sum_{i=1}^{N}\mathcal{I}_{M}\frac{1}{K-1}\sum_{k\neq y_i}\left(\frac{1}{K-1}\sum_{k\neq y_i}\sigma\left(\mathcal{M}(\mathbf{x}_i)\right)_k - \sigma\left(\mathcal{M}(\mathbf{x}_i)\right)_{y_i}\right)^2\\
&\stackrel{(b)}{\approx}\frac{1}{N^M}\sum_{i=1}^{N}\mathcal{I}_{M}\frac{1}{K-1}\sum_{k\neq y_i}\left(\frac{1}{K-1}\sum_{k\neq y_i}\sigma\left(\mathcal{M}(\mathbf{x}_i)\right)_k - \sigma\left(\mathcal{M}(\mathbf{x}_i)\right)_{k}\right)^2\\
\end{split}
\end{align}
Given the convexity of the mean square loss function, the inference of (a) can be efficiently deduced using Jensen's inequality. Moreover, when considering misclassified samples, assuming a uniform distribution of logits for correct labels is plausible relative to other logits. This assumption enables us to approximate $\sigma\left(\mathcal{M}(\mathbf{x}_i)\right){y_i}$ with $\sigma\left(\mathcal{M}(\mathbf{x}_i)\right){k}$. Therefore, under these premises, the validity of (b) is affirmed, and the expression on its right side can be represented as $\check{\mathcal{P}}_{M}^{N}$.

Throughout the implementation of the SLORE procedure, effective regularization of $\mathcal{P}_{C}^N$ and $\mathcal{P}_{M}^N$ is achieved by concentrating on their respective lower bounds, $\check{\mathcal{P}}_{C}^N$ and $\check{\mathcal{P}}_{M}^N$. This approach facilitates the efficient optimization of the lower bounds $\overline{\gamma}_{ce}^{N^M}$ and $\overline{\gamma}_{ce}^{N^C}$ in relation to $\hat{\gamma}_{ce}^{N^{M}}$and $\hat{\gamma}_{ce}^{N^{C}}$.

\subsubsection{Adaptive Adversarial Logit Pairing}
The surrogate terms $\check{\mathcal{P}}_{M}^N$ and $\check{\mathcal{P}}_{C}^N$ serve as an innovative method for regularizing $\hat{\gamma}_{ce}^{N^M}$ and $\hat{\gamma}_{ce}^{N^C}$. However, as explicated in Eq.\ref{regularize_correct_bound} and Eq.\ref{regularize_wrong_bound}, penalizing $\mathcal{P}_{M}^N$ and $\mathcal{P}_{C}^N$ may inadvertently lead to the standard logits being paired with logits from an unknown distribution. This potential misalignment poses a risk to the model's overall robustness.

\begin{algorithm}
\caption{Logit-Oriented Adversarial Training (LOAT)}
\label{alg}
\scriptsize
\begin{algorithmic}[1]
    \State \textbf{Input:} Victim model $\mathcal{M}$, training set ($X$, $Y$), $\mathcal{A}_{p}^{\epsilon}(\cdot)$-based adversarial training objective $\mathcal{L}_{\mathcal{A}_{p}^{\epsilon}}(\cdot)$, epoch-wise breakpoints $\mathcal{E}_{1}$ and $\mathcal{E}_{2}$, number of training epochs $\hat{\mathcal{E}}$, hyperparameter $\tau$ and $\gamma$, LOAT type $T$, and updated loss $\mathsf{L}$.
    \State \textbf{Output:} LOAT-boosted adversarial-trained model $\mathcal{M}$.
    \For{$\mathcal{E}$ = 1, 2, ..., $\hat{\mathcal{E}}$}
        \For{($X^{j}$, $Y^{j}$) in batch $j$}
        \If{$\mathcal{E}\leq\mathcal{E}_1$}
                \If{$T$ = SLORE}
                    \State 
                $\mathsf{L}$ = $\mathcal{L}_{\mathcal{A}_{p}^{\epsilon}}((X^j, Y^j), \mathcal{M})$ + $\tau\cdot(\check{\mathcal{P}}_{C}^{N_j}-\check{\mathcal{P}}_{M}^{N_j})$
                \ElsIf{$T$=LORE}
                    \State
                     $\mathsf{L}$ = $\mathcal{L}_{\mathcal{A}_{p}^{\epsilon}}((X^j, Y^j), \mathcal{M})$ + $\tau\cdot(\check{\mathcal{P}}_{C}^{N_j}-\check{\mathcal{P}}_{M}^{N_j})$ + $\gamma*\mathcal{AP}_{M}^{N_j}$
                \EndIf
                
            \ElsIf{$\mathcal{E}\geq\mathcal{E}_2$}
                \If{$T$ = SLORE}
                    \State 
                $\mathsf{L}$ = $\mathcal{L}_{\mathcal{A}_{p}^{\epsilon}}((X^j, Y^j), \mathcal{M})$ + $\tau\cdot(\check{\mathcal{P}}_{M}^{N_j}-\check{\mathcal{P}}_{C}^{N_j})$
                \ElsIf{$T$=LORE}
                    \State
                     $\mathsf{L}$ = $\mathcal{L}_{\mathcal{A}_{p}^{\epsilon}}((X^j, Y^j), \mathcal{M})$ + $\tau\cdot(\check{\mathcal{P}}_{M}^{N_j}-\check{\mathcal{P}}_{C}^{N_j})$ + $\gamma*\mathcal{AP}_{C}^{N_j}$
                \EndIf
            \Else
                \State 
                $\mathsf{L}$ = $\mathcal{L}_{\mathcal{A}_{p}^{\epsilon}}((X^i, Y^i), \mathcal{M})$
            \EndIf
\State $\mathcal{M} \gets$ Update Model With Loss $\mathsf{L}$.
        \EndFor
    \EndFor
    \State \Return $\mathcal{M}$
\end{algorithmic}
\end{algorithm}
To mitigate the potential drawbacks identified, we incorporate Adversarial Logit Pairing (ALP) as described in~\cite{kannan2018adversarial}, tailoring an adversary-focused regularization strategy. In the early stages of training, where standard penalization is applied for the decrease of $\check{\mathcal{P}}^N_M$, we introduce an adversarial penalty, $\mathcal{AP}^N_M$. This penalty is designed to diminish the disparity between the logits of misclassified clean samples and their adversarial counterparts. In contrast, during the latter epochs, we apply a similar adversarial penalty, $\mathcal{AP}^N_C$, but focus on reducing the logit distance for correctly classified clean samples compared to their adversarial equivalents. By employing the adversarial generator $\mathcal{A}_{p}^{\epsilon}(\cdot)$, we can formally express $\mathcal{AP}^N_M$ and $\mathcal{AP}^N_C$ following the derivations of $\check{\mathcal{P}}_{M}^N$ and $\check{\mathcal{P}}_{C}^N$:
{
\small
\begin{align}
\begin{split}
\mathcal{AP}^{N}_{M} &= \frac{1}{N^M}\sum_{i=1}^{N}\mathcal{I}_{M}\sum_{k=1}^{K}(\sigma(\mathcal{M}(\mathbf{x}_i))_k-\sigma(\mathcal{A}_{p}^{\epsilon}(\mathcal{M}(\mathbf{x}_i)))_k)^2 \\
\mathcal{AP}^{N}_{C} &= \frac{1}{N^C}\sum_{i=1}^{N}\mathcal{I}_{C}\sum_{k=1}^{K}(\sigma(\mathcal{M}(\mathbf{x}_i))_k-\sigma(\mathcal{A}_{p}^{\epsilon}(\mathcal{M}(\mathbf{x}_i)))_k)^2
\end{split}
\end{align}
}
More precisely, the integration of standard logit-oriented regularization with adaptive adversarial logit pairing forms the basis of what we call Logit-Oriented REgularization (LORE). Comprehensive details and the procedural steps of the corresponding training algorithm are delineated in Alg.~\ref{alg}.
\begin{table}
\scriptsize
\setlength{\tabcolsep}{0.9pt}
\centering
\caption{Classification Accuracy on Cifar10 (\%). \textbf{Bold} and \textit{italic} texts represent the highest and the second-highest accuracy in each block respectively.}
\begin{tabular}{c|c|c|c|c|c|c|c}
\toprule
Models & Defense & Clean$_{tr}$ & Clean$_{te}$ & FGSM & PGD$^{7}$ & PGD$^{20}$ & T/E \\
\midrule
\multirow{14}{*}{ResNet18}   & TRADES & 93.39 & 82.23 & 57.89  & \textit{{53.68}}  & \textbf{{52.00}} & 216s\\
                            & TRADES+SLORE & \textit{94.77}&\textit{{82.29}} & \textit{{62.19}} & 53.32 & 49.77
                            &
                            218s
                           \\
                            &TRADES+LORE& \textbf{95.35} &  \textbf{{82.51}} &\textbf{{62.47}} & \textbf{{53.96}} & \textit{{50.43}}&219s  \\
                            \cmidrule(lr){2-8}
                             & TRADES(S) & \textit{{92.92}} & \textit{{82.95}}& 58.14 & 54.01 & \textbf{{52.71}} &206s\\
                             & TRADES(S)+SLORE & \textbf{{93.25}} & \textbf{{83.01}} & \textit{{62.32}} & \textbf{{54.99}} & \textit{{51.96}}
                             &208s\\
                         & TRADES(S)+LORE& 92.79 & 82.73 & \textbf{{62.76}} & \textit{{54.06}} & 51.03 & 209s\\    \cmidrule(lr){2-8}
                             & MART & 87.85 & 80.61 & 62.41 & 54.86 & 51.65
                             & 98s\\
                             & MART+SLORE &\textit{{88.07}}& \textit{{80.81}} & \textit{{62.61}} &\textit{{55.06}}& \textit{{51.89}}& 100s \\
                             & MART+LORE & \textbf{{89.04}} & \textbf{{81.53}} & \textbf{{63.35}} &\textbf{{55.58}}& \textbf{{52.15}}
                             & 100s\\
                             
                             
                             \cmidrule(lr){2-8}
                             & PGD-AT & \textbf{{97.62}} & 83.19 & 59.99 & 49.86 &44.94& 158s\\
                             & PGD-AT+SLORE & 97.13& \textbf{{83.41}} & \textbf{{61.80}} &\textbf{{51.46}} &\textbf{{46.50}}& 160s \\
                             & PGD-AT+LORE & \textit{{97.38}}& \textit{{83.25}} & \textit{{60.94}} & \textit{{50.28}} &\textit{{45.30}}& 161s\\
\midrule
\multirow{8}{*}{\centering ResNet50}  
                             & TRADES & 93.89  &
                             83.38&  \textit{{64.53}} & \textit{{56.23}} & \textbf{{52.39}} & 674s\\ 
                             & TRADES+SLORE & \textit{{94.74}} & \textbf{{84.07}} & 64.32 &55.78 & 52.15 & 676s \\
                             & TRADES+LORE & \textbf{{94.83}} & \textit{{83.81}} & \textbf{{64.69}} &\textbf{{56.30}} & \textit{{52.34}} & 679s \\ \cmidrule(lr){2-8}
                            & MART & 87.66 & 80.49 & \textit{{62.66}} & \textit{{55.19}} & \textit{{51.62}} & 209s\\
                            & MART+SLORE & \textit{{88.25}}  & \textbf{{81.29}} & 62.61 & 54.76 &51.23 & 213s\\
                            &MART+LORE & \textbf{{88.29}}  & \textit{{80.54}} & \textbf{{62.90}} & \textbf{{55.43}} &\textbf{{52.00}} & 214s\\
                            \cmidrule(lr){2-8}
                            & PGD-AT & \textit{94.77}& 82.92 &
                        60.79& 50.74&46.03&174s\\
                             & PGD-AT+SLORE &93.51& \textit{{82.92}} & \textit{{62.34}}& \textit{{52.44}} & 48.14 & 184s \\
& PGD-AT+LORE &\textbf{{95.09}}& \textbf{{83.05}} & \textbf{{62.94}}& \textbf{{53.37}} & \textbf{{48.98}} & 185s \\
\midrule
\multirow{8}{*}{\centering WRN-34-10}  &TRADES & 98.98  & 84.57 & \textit{{63.45}}  &52.60 & \textbf{{48.15}} & 755s\\ 
                             & TRADES+SLORE &99.65 & \textbf{{85.22}}& 63.34& \textit{{52.66}}&47.60 & 763s\\
                            & TRADES+LORE &\textbf{{99.69}} & \textit{{85.14}}& \textbf{{63.49}}& \textbf{{52.89}}&\textit{{47.84}}& 768s \\
                            \cmidrule(lr){2-8}
                            & MART & 90.21 & 84.36 & 68.26 & 71.55 & 58.34 & 369s\\
                            & MART+SLORE&\textbf{{91.11}} & \textbf{{85.17}} & \textbf{{69.35}} & \textbf{{72.87}}&\textbf{{59.10}} & 374s\\
                         & MART+LORE&\textit{{90.89}} & \textit{{85.00}} & \textit{{68.92}} & \textit{{72.32}}&\textit{{59.64}} & 374s\\
            \cmidrule(lr){2-8}
                            & PGD-AT & \textbf{{99.90}}& 84.47 &59.89 & 48.48 & 43.68 & 520s\\
                             & PGD-AT+SLORE & \textit{{99.88}} & \textit{{84.68}} & \textit{{60.18}} & \textit{{49.49}} & \textit{{44.79}} & 528s\\
                             & PGD-AT+LORE &99.81 & \textbf{{85.29}} & \textbf{{61.04}} & \textbf{{49.53}} & \textbf{{44.79}} & 529s\\
\bottomrule
\end{tabular}
\label{tab_traditional}
\end{table}

\section{Experiments}
\label{sec:experiments}
In this section, we comprehensively evaluate the performance enhancement of LOAT regarding standard accuracy and adversarial robustness. We test against a spectrum of attacks, including FGSM~\cite{goodfellow2014explaining}, PGD-7/20/40~\cite{madry2017towards}, and AutoAttack (AA)~\cite{croce2020reliable}.

\textbf{Experimental Setup.} Our experiments are conducted under the $\ell_\infty$ threat model. We set the perturbation radius ($\epsilon$) to 8/255, the number of perturbation steps ($k$) to 10, and the step size ($\alpha$) to 2/255. All experiments utilize dual GeForce RTX 3090 GPUs, with each experimental run replicated thrice to ensure reliability and consistency.

For the CIFAR10 dataset, our models include ResNet-18, ResNet-50, and WideResNet-34-10, each employing PGD-AT, TRADES, TRADES (LSE) (also referred to as TRADES (S)), and MART, respectively. The trade-off factor ($\beta$) for TRADES, TRADES (LSE), and MART is uniformly set at 6.0.
The training duration is set for 110 epochs. As outlined in Alg.~\ref{alg}, we establish epoch-wise breakpoints, $\mathcal{E}_1$ and $\mathcal{E}_2$, at 1 and 100 for ResNet-18 and ResNet-50, and 1 and 85 for WideResNet-34-10, respectively.

For experiments on the Elucidating Diffusion Model-augmented CIFAR10 (DM-AT), we use ResNet-18 and WideResNet-34-10 as the backbones, conducting TRADES, TRADES (S), and MART individually. Here, $\beta$ is set at 5.0, with the training spanning 100 epochs. The breakpoints $\mathcal{E}_1$ and $\mathcal{E}_2$ are set at 1 and 90, respectively.
\begin{table}
\scriptsize
\setlength{\tabcolsep}{2pt}
\centering
\caption{Classification Accuracy of models trained by 1$\times 10^6$ EDM-generated images-augmented Cifar10 (\%).}
\begin{tabular}{c|c|c|c|c|c|c}
\toprule
Models & Defense & Clean$_{tr}$ & Clean$_{te}$ & PGD$^{40}$  & AA & T/E \\
\midrule
\multirow{6}{*}{\centering\makecell{PreResNet18 \\ (Swish)}}   & TRADES & 92.54 & 86.17  & 58.47 & \textbf{{54.65}} & 744s \\
                              & TRADES+SLORE
                              & \textit{{94.48}}
                              & \textit{{88.35}} & \textit{{59.00}} & 53.76 &749s\\
                              & TRADES+LORE
                              & \textbf{{94.69}}
                              & \textbf{{88.46}} & \textbf{{59.00}} & \textit{{53.87}} &750s\\
                 \cmidrule(lr){2-7}
                             & TRADES(S)
                             &
                             92.64
                             & 86.52  & \textit{{59.74}} & \textit{{54.61}} & 743s\\
                             & TRADES(S)+SLORE 
                             &
                            \textbf{{94.91}}
                             & \textbf{{88.94}} & 58.87 & 52.41 & 745s\\
& TRADES(S)+LORE 
                             &
                             \textit{{93.30}}
                             & \textit{{86.85}} & \textbf{{60.08}} & \textbf{{55.06}} & 745s\\
                 \cmidrule(lr){2-7}
                             & MART 
                             &
                             85.52
                             & 84.86   & 58.51 & 51.70 & 745s\\
                             & MART+SLORE 
                             &
                             \textbf{{89.25}}
                             & \textit{{86.29}}  &\textbf{{62.11}}& \textbf{{51.96}} & 747s\\
& MART+LORE 
                             &
                             \textit{{87.75}}
                             &\textbf{{87.05}}
                             &\textit{{61.67}}& \textit{{51.92}} & 747s\\
                             
\midrule
\multirow{6}{*}{\centering\makecell{WRN-28-10 \\ (Swish)}}  &TRADES & 94.80 & 88.87   & \textit{{63.16}} & \textbf{{59.44}} & 754s\\ 
                             & TRADES+SLORE 
                             & \textit{{96.39}}&\textit{{90.36}}  & 62.95 & 58.12 & 756s \\
                             & TRADES+LORE 
                             & \textbf{{96.58}}&\textbf{{90.67}}  & \textbf{{63.57}} & \textit{{58.74}} & 757s \\
                 \cmidrule(lr){2-7}
                            & TRADES(S) 
                            & 94.76
                            & 88.66   & \textit{{63.78}} & \textit{{58.85}} & 784s\\
                            & TRADES(S)+SLORE
                            &\textbf{{96.40}} &\textbf{{90.82}} & 63.09 & 56.86& 798s
                            \\
                            & TRADES(S)+LORE
                            &\textit{{95.30}} &\textit{{89.45}} & \textbf{{64.17}} & \textbf{{59.54}} & 798s
                            \\
                 \cmidrule(lr){2-7} 
                            & MART 
                            & 87.22
                            & 88.83  & 62.91 & 57.08 & 755s\\
                             & MART+SLORE 
                             & \textbf{{90.50}}
                             & \textit{{89.82}}  & \textit{{65.78}} & \textit{{57.44}} & 757s\\
& MART+LORE
                            & \textit{{89.75}}
                            & \textbf{{90.18}}  & \textbf{{66.03}} & \textbf{{58.07}} & 758s\\
                            
\bottomrule
\end{tabular}
\label{tab_edm}
\end{table}
\textbf{Boost in Standard Accuracy.} As detailed in Tab.~\ref{tab_traditional}, MART+LORE achieves a significant 0.92\% increase in accuracy for ResNet-18. For ResNet-50 and WideResNet-34-10, MART+SLORE leads to improvements of 0.80\% and 0.81\%, respectively. In the case of TRADES+SLORE, there is an enhancement in clean accuracy by 0.69\% for ResNet-50 and 0.65\% for WideResNet-34-10. Additionally, PGD-AT+LORE boosts WideResNet-34-10 performance by 0.82\%.

These performance gains are even more pronounced in settings augmented with EDM-generated data. As Tab.~\ref{tab_edm} indicates, TRADES+LORE registers a substantial increase of 2.29\% and 1.80\% in standard accuracy for Swish-activated PreAct ResNet-18 (abbreviated as PreResNet18 (Swish)) and Swish-activated WideResNet-28-10 (abbreviated as WRN-28-10 (Swish)), respectively. Meanwhile, MART achieves boosts of 2.19\% and 1.35\% in these models. Further, TRADES (S)+SLORE records increases of 2.42\% and 2.16\% in PreResNet18 (Swish) and WRN-28-10 (Swish), respectively.

\textbf{Boost in Adversarial Accuracy.} In the case of CIFAR10 trained models (referenced in Tab.~\ref{tab_traditional}), LOAT significantly bolsters the robustness of MART and PGD-AT against all evaluated adversarial attacks. Although its impact on TRADES and TRADES (S) is less pronounced, it still ensures effective defence, especially in ResNet-18. For instance, TRADES's accuracy against the FGSM attack is improved by 4.58\%, and TRADES (S)'s by 4.62\%. In scenarios involving DM-AT, our defence strategy demonstrates even greater effectiveness. Notably, against the highly potent AutoAttack, our approach distinctly improves the robustness of TRADES (S) and MART. Specifically, the accuracy of TRADES (S) against AutoAttack is increased by 0.45\% for PreResNet18 (Swish) and by 0.69\% for WideResNet-28-10 (Swish). Similarly, MART's accuracy against AutoAttack is elevated by 0.26\% for PreResNet18 (Swish) and 0.99\% for WideResNet-28-10 (Swish).

\textbf{Ablation Study.} In accordance with Alg.~\ref{alg}, we assess the effectiveness of applying regularizations before $\mathcal{E}_1$, after $\mathcal{E}_2$, as well as their combination.  Additionally, we examine the impacts of regularizations of $\check{\mathcal{P}}_{C}$ and $\check{\mathcal{P}}_{M}$ individually and in combination. Detailed experimental results are shown in Tab.~\ref{tab:combined_ablation_study_epoch},~\ref{tab:combined_ablation_study_penalty}. Our findings indicate that for MART, which has been regularized, the sole application of $\check{\mathcal{P}}_{M}$ yields the most favourable results. We hypothesize that this is due to MART's already robust standard generalization performance, which renders the influence of $\check{\mathcal{P}}_{C}$ less pronounced.
\begin{table}
\centering
\caption{Ablation study for $\leftarrow\mathcal{E}_1$ and $\mathcal{E}_2\rightarrow$ on ResNet-18 (\%) .}
\resizebox{\linewidth}{!}{
\begin{tabular}{l|ccccc|ccccc}
\hline
& \multicolumn{5}{c|}{SLORE} & \multicolumn{5}{c}{LORE} \\
\cline{2-11}
PGD-AT &Clean$_{tr}$ &Clean$_{te}$ & FGSM & PGD{\tiny$^{7}$} & PGD{\tiny$^{20}$} &Clean$_{tr}$ & Clean$_{te}$ & FGSM & PGD{\tiny$^{7}$} & PGD{\tiny$^{20}$} \\
\hline
$\leftarrow\mathcal{E}_1$& 96.84 & 83.39 & 59.80 & 48.31 & 43.86 & 96.50 & 83.07 & 59.75 & 48.80 & 44.09 \\
$\mathcal{E}_2\rightarrow$ & 96.34 & 82.65 & 59.68 & 49.33 & 45.18 & 96.65 & 82.67 & 60.46 & \textbf{50.85} & \textbf{45.54}  \\
$\leftarrow\mathcal{E}_1$, $\mathcal{E}_2\rightarrow$ & \textbf{97.13} & \textbf{83.41} & \textbf{61.80} & \textbf{51.46} & \textbf{46.50} & \textbf{97.38} & \textbf{83.25} & \textbf{60.94} & 50.28 & 45.30 \\
\hline
\end{tabular}}

\vspace{1mm} 
\resizebox{\linewidth}{!}{
\begin{tabular}{l|ccccc|ccccc}
\hline
& \multicolumn{5}{c|}{SLORE} & \multicolumn{5}{c}{LORE} \\
\cline{2-11}
MART & Clean$_{tr}$ & Clean$_{te}$ & FGSM & PGD{\tiny$^{7}$} & PGD{\tiny$^{20}$} & Clean$_{tr}$ & Clean$_{te}$ & FGSM & PGD{\tiny$^{7}$} & PGD{\tiny$^{20}$} \\
\hline
$\leftarrow\mathcal{E}_1$ & 88.03 & 80.76 & 62.60 & \textbf{55.18} & 51.64 & 88.01 & 80.62 & 62.81 & 55.25 & 51.75 \\
$\mathcal{E}_2\rightarrow$ & 88.04 & 80.79 & 62.58 & 55.13 & 51.67 & 88.28 & 80.89 & 62.92 & 55.60 & \textbf{52.57} \\
$\leftarrow\mathcal{E}_1$, $\mathcal{E}_2\rightarrow$ & \textbf{88.07} & \textbf{80.81} & \textbf{62.61} & 55.06 & \textbf{51.89} & \textbf{89.04} & \textbf{81.53} & \textbf{63.35} & \textbf{55.58} & 52.15 \\
\hline
\end{tabular}}
\label{tab:combined_ablation_study_epoch}
\end{table}

\begin{table}
\centering
\caption{Ablation study for $\check{\mathcal{P}}_{C}$ and $\check{\mathcal{P}}_{M}$ on ResNet-18 (\%).}
\resizebox{\linewidth}{!}{
\begin{tabular}{l|ccccc|ccccc}
\hline
& \multicolumn{5}{c|}{SLORE} & \multicolumn{5}{c}{LORE} \\
\cline{2-11}
PGD-AT &Clean$_{tr}$ &Clean$_{te}$ & FGSM & PGD{\tiny$^{7}$} & PGD{\tiny$^{20}$} &Clean$_{tr}$ & Clean$_{te}$ & FGSM & PGD{\tiny$^{7}$} & PGD$^{20}$ \\
\hline
$\check{\mathcal{P}}_{C}$& 96.49 & 83.01 & 59.71 & 49.62 & 44.71 & 96.15 & 82.80 & 60.02 & 49.40 & 44.77 \\
$\check{\mathcal{P}}_{M}$ & 96.78 & 82.92 & 60.79 & 50.19 & 45.52 & 97.09 & 82.72 & 60.19 & 49.68 & \textbf{45.44}  \\
$\check{\mathcal{P}}_{C}$, $\check{\mathcal{P}}_{M}$ & \textbf{97.13} & \textbf{83.41} & \textbf{61.80} & \textbf{51.46} & \textbf{46.50} & \textbf{97.38} & \textbf{83.25} & \textbf{60.94} & \textbf{50.28} & 45.30 \\
\hline
\end{tabular}}

\vspace{1mm} 

\resizebox{\linewidth}{!}{
\begin{tabular}{l|ccccc|ccccc}
\hline
& \multicolumn{5}{c|}{SLORE} & \multicolumn{5}{c}{LORE} \\
\cline{2-11}
MART & Clean$_{tr}$ & Clean$_{te}$ & FGSM & PGD{\tiny$^{7}$} & PGD{\tiny$^{20}$} & Clean$_{tr}$ & Clean$_{te}$ & FGSM & PGD{\tiny$^{7}$} & PGD{\tiny$^{20}$} \\
\hline
$\check{\mathcal{P}}_{C}$ & 87.98 & 80.22 & 62.77 & 55.29 & 52.24 & 88.01 & 80.62 & 62.81 & 55.25 & 51.75 \\
$\check{\mathcal{P}}_{M}$ & \textbf{88.34} & \textbf{80.88} & \textbf{62.99} & \textbf{55.24} & \textbf{51.94} & 88.28 & 80.89 & 62.92 & \textbf{55.60} & \textbf{52.57} \\
$\check{\mathcal{P}}_{C}$, $\check{\mathcal{P}}_{M}$ & 88.07 & 80.81 & 62.61 & 55.06 & 51.89 & \textbf{89.04} & \textbf{81.53} & \textbf{63.35} & 55.58 & 52.15 \\
\hline
\end{tabular}}
\label{tab:combined_ablation_study_penalty}
\end{table}
\begin{table}[ht]
\centering
\caption{Evaluation of LOAT-boosted S2O on ResNet-18 (\%).}
\scriptsize 
\setlength{\tabcolsep}{0.5pt}
\begin{tabular}{l|cccc|cccc|cccc}
\hline
\multirow{2}{*}{S2O} & \multicolumn{4}{c|}{TRADES} & \multicolumn{4}{c|}{MART} & \multicolumn{4}{c}{PGD-AT} \\
\cline{2-13}
& \tiny Clean$_{te}$ & \tiny PGD$^7$ & \tiny PGD$^{20}$ & \tiny PGD$^{40}$ & \tiny Clean$_{te}$ & \tiny PGD$^7$ & \tiny PGD$^{20}$ & \tiny PGD$^{40}$& \tiny Clean$_{te}$ & \tiny PGD$^7$ & \tiny PGD$^{20}$ & \tiny PGD$^{40}$ \\
\hline
Vanilla & 83.90 & 54.87 & 52.11 & 51.98 & 82.08 & 54.84 & \textit{{51.66}} & \textit{{51.38}} & 84.41 & 50.98 & 47.05 & 46.57 \\
SLORE & \textit{{84.30}} & \textit{{55.06}} & \textbf{{52.31}} & \textit{{52.05}} & \textbf{{82.45}} & \textit{{54.84}} & 51.53 & 51.23 & \textbf{{84.90}} & \textbf{{52.39}} & \textbf{{47.92}} & \textbf{{47.54}} \\
LORE & \textbf{{84.90}} & \textbf{{55.08}} & \textit{{52.26}} & \textbf{{52.19}} & \textit{{82.32}} & \textbf{{55.05}} & \textbf{{52.37}} & \textbf{{52.14}} & \textit{{84.86}} & \textit{{51.31}} & \textit{{47.24}} & \textit{{46.74}}\\
\hline
\end{tabular}
\label{tab:s2o}
\end{table}

\textbf{Adaptibility to Weight Regularizations.} LOAT is architecture-agnostic and algorithm-agnostic, suggesting that it can be integrated seamlessly into other weight-oriented regularization algorithms to enhance adversarial training. To empirically validate this hypothesis, we apply our framework to train algorithms boosted by Second-Order Statistics Optimization (S2O). The extensive results of this integration are displayed in Tab.~\ref{tab:s2o}.

\textbf{Time Efficiency.} We measure the time taken for each epoch (Time/Epochs) and present the detailed findings in Tab.~\ref{tab_traditional} and~\ref{tab_edm}, which confirm that LOAT introduces minimal computational overhead, maintaining time efficiency.

\textbf{More Experiments.} To solidify the credibility of LOAT, we extended our experimentation to CIFAR100 and SVHN. Additional results can be found in the Appendix.

\section{Conclusions}
\label{sec:conclusions}
In this paper, we address the challenge of maintaining standard generalization performance in adversarial-trained deep neural networks. We explore the Rademacher complexity of ReLU-activated MLPs through the lens of the Fisher-Rao norm. This perspective enables us to introduce a logit-based variable, which exhibits sensitivity to various factors related to model complexity, including model width and training objective. Also, it shows a strong correlation with the generalization gap of CE loss between adversarial-trained and standard-trained models. We leverage these insights to develop the Logit-Oriented Adversarial Training (LOAT) framework, enhancing adversarial training without significant computational cost. Our comprehensive experiments on prevalent adversarial training algorithms and diverse network architectures confirm the effectiveness of LOAT in mitigating the trade-off between robustness and accuracy.
\section{Acknowledgements}
This work is supported by the University of Liverpool and the China Scholarship Council (CSC).
{
    \small
    \bibliographystyle{ieeenat_fullname}
    \bibliography{main}

\begin{thebibliography}{43}
\providecommand{\natexlab}[1]{#1}
\providecommand{\url}[1]{\texttt{#1}}
\expandafter\ifx\csname urlstyle\endcsname\relax
  \providecommand{\doi}[1]{doi: #1}\else
  \providecommand{\doi}{doi: \begingroup \urlstyle{rm}\Url}\fi

\bibitem[Bartlett et~al.(2017)Bartlett, Foster, and Telgarsky]{bartlett2017spectrallynormalized}
Peter Bartlett, Dylan~J. Foster, and Matus Telgarsky.
\newblock Spectrally-normalized margin bounds for neural networks, 2017.

\bibitem[Bartlett and Maass(2003)]{bartlett2003vapnik}
Peter~L Bartlett and Wolfgang Maass.
\newblock Vapnik-chervonenkis dimension of neural nets.
\newblock \emph{The handbook of brain theory and neural networks}, pages 1188--1192, 2003.

\bibitem[Bartlett and Mendelson(2002)]{bartlett2002rademacher}
Peter~L Bartlett and Shahar Mendelson.
\newblock Rademacher and gaussian complexities: Risk bounds and structural results.
\newblock \emph{Journal of Machine Learning Research}, 3\penalty0 (Nov):\penalty0 463--482, 2002.

\bibitem[Bojarski et~al.(2016)Bojarski, Testa, Dworakowski, Firner, Flepp, Goyal, Jackel, Monfort, Muller, Zhang, Zhang, Zhao, and Zieba]{self-driving}
Mariusz Bojarski, Davide~Del Testa, Daniel Dworakowski, Bernhard Firner, Beat Flepp, Prasoon Goyal, Lawrence~D. Jackel, Mathew Monfort, Urs Muller, Jiakai Zhang, Xin Zhang, Jake Zhao, and Karol Zieba.
\newblock End to end learning for self-driving cars, 2016.

\bibitem[Croce and Hein(2020)]{croce2020reliable}
Francesco Croce and Matthias Hein.
\newblock Reliable evaluation of adversarial robustness with an ensemble of diverse parameter-free attacks.
\newblock In \emph{International conference on machine learning}, pages 2206--2216. PMLR, 2020.

\bibitem[Goodfellow et~al.(2014)Goodfellow, Shlens, and Szegedy]{goodfellow2014explaining}
Ian~J Goodfellow, Jonathon Shlens, and Christian Szegedy.
\newblock Explaining and harnessing adversarial examples.
\newblock \emph{arXiv preprint arXiv:1412.6572}, 2014.

\bibitem[Grosse et~al.(2017)Grosse, Papernot, Manoharan, Backes, and McDaniel]{malware}
Kathrin Grosse, Nicolas Papernot, Praveen Manoharan, Michael Backes, and Patrick McDaniel.
\newblock Adversarial examples for malware detection.
\newblock In \emph{Computer Security--ESORICS 2017: 22nd European Symposium on Research in Computer Security, Oslo, Norway, September 11-15, 2017, Proceedings, Part II 22}, pages 62--79. Springer, 2017.

\bibitem[Hu et~al.(2021)Hu, Chu, Pei, Liu, and Bian]{hu2021model}
Xia Hu, Lingyang Chu, Jian Pei, Weiqing Liu, and Jiang Bian.
\newblock Model complexity of deep learning: A survey.
\newblock \emph{Knowledge and Information Systems}, 63:\penalty0 2585--2619, 2021.

\bibitem[Huang et~al.(2020)Huang, Kroening, Ruan, Sharp, Sun, Thamo, Wu, and Yi]{huang2020survey}
Xiaowei Huang, Daniel Kroening, Wenjie Ruan, James Sharp, Youcheng Sun, Emese Thamo, Min Wu, and Xinping Yi.
\newblock A survey of safety and trustworthiness of deep neural networks: Verification, testing, adversarial attack and defence, and interpretability.
\newblock \emph{Computer Science Review}, 37:\penalty0 100270, 2020.

\bibitem[Huang et~al.(2023{\natexlab{a}})Huang, Jin, and Ruan]{huang2012deep}
Xiaowei Huang, Gaojie Jin, and Wenjie Ruan.
\newblock Deep reinforcement learning.
\newblock In \emph{Machine Learning Safety}, pages 219--235. Springer, 2023{\natexlab{a}}.

\bibitem[Huang et~al.(2023{\natexlab{b}})Huang, Ruan, Huang, Jin, Dong, Wu, Bensalem, Mu, Qi, Zhao, et~al.]{huang2023survey}
Xiaowei Huang, Wenjie Ruan, Wei Huang, Gaojie Jin, Yi Dong, Changshun Wu, Saddek Bensalem, Ronghui Mu, Yi Qi, Xingyu Zhao, et~al.
\newblock A survey of safety and trustworthiness of large language models through the lens of verification and validation.
\newblock \emph{arXiv preprint arXiv:2305.11391}, 2023{\natexlab{b}}.

\bibitem[Jin et~al.(2022)Jin, Yi, Huang, Schewe, and Huang]{jin2022enhancing}
Gaojie Jin, Xinping Yi, Wei Huang, Sven Schewe, and Xiaowei Huang.
\newblock Enhancing adversarial training with second-order statistics of weights, 2022.

\bibitem[Kannan et~al.(2018)Kannan, Kurakin, and Goodfellow]{kannan2018adversarial}
Harini Kannan, Alexey Kurakin, and Ian Goodfellow.
\newblock Adversarial logit pairing, 2018.

\bibitem[Krogh and Hertz(1991)]{krogh1991simple}
Anders Krogh and John Hertz.
\newblock A simple weight decay can improve generalization.
\newblock \emph{Advances in neural information processing systems}, 4, 1991.

\bibitem[Liang et~al.(2019)Liang, Poggio, Rakhlin, and Stokes]{liang2019fisher}
Tengyuan Liang, Tomaso Poggio, Alexander Rakhlin, and James Stokes.
\newblock Fisher-rao metric, geometry, and complexity of neural networks.
\newblock In \emph{The 22nd international conference on artificial intelligence and statistics}, pages 888--896. PMLR, 2019.

\bibitem[Lyu et~al.(2015)Lyu, Huang, and Liang]{lyu2015unified}
Chunchuan Lyu, Kaizhu Huang, and Hai-Ning Liang.
\newblock A unified gradient regularization family for adversarial examples, 2015.

\bibitem[Madry et~al.(2017)Madry, Makelov, Schmidt, Tsipras, and Vladu]{madry2017towards}
Aleksander Madry, Aleksandar Makelov, Ludwig Schmidt, Dimitris Tsipras, and Adrian Vladu.
\newblock Towards deep learning models resistant to adversarial attacks.
\newblock \emph{arXiv preprint arXiv:1706.06083}, 2017.

\bibitem[Moosavi-Dezfooli et~al.(2018)Moosavi-Dezfooli, Fawzi, Uesato, and Frossard]{moosavidezfooli2018robustness}
Seyed-Mohsen Moosavi-Dezfooli, Alhussein Fawzi, Jonathan Uesato, and Pascal Frossard.
\newblock Robustness via curvature regularization, and vice versa, 2018.

\bibitem[Mu et~al.(2021)Mu, Ruan, Soriano~Marcolino, and Ni]{mu2021sparse}
Ronghui Mu, Wenjie Ruan, Leandro Soriano~Marcolino, and Qiang Ni.
\newblock Sparse adversarial video attacks with spatial transformations.
\newblock In \emph{The 32nd British Machine Vision Conference (BMVC'21)}, 2021.

\bibitem[Mu et~al.(2023)Mu, Ruan, Marcolino, Jin, and Ni]{mu2023certified}
Ronghui Mu, Wenjie Ruan, Leandro~Soriano Marcolino, Gaojie Jin, and Qiang Ni.
\newblock Certified policy smoothing for cooperative multi-agent reinforcement learning.
\newblock In \emph{Proceedings of the AAAI Conference on Artificial Intelligence (AAAI'23)}, 2023.

\bibitem[Nakkiran(2019)]{tradeoff-nak}
Preetum Nakkiran.
\newblock Adversarial robustness may be at odds with simplicity, 2019.

\bibitem[Neyshabur et~al.(2015{\natexlab{a}})Neyshabur, Salakhutdinov, and Srebro]{neyshabur2015pathsgd}
Behnam Neyshabur, Ruslan Salakhutdinov, and Nathan Srebro.
\newblock Path-sgd: Path-normalized optimization in deep neural networks, 2015{\natexlab{a}}.

\bibitem[Neyshabur et~al.(2015{\natexlab{b}})Neyshabur, Tomioka, and Srebro]{neyshabur2015normbased}
Behnam Neyshabur, Ryota Tomioka, and Nathan Srebro.
\newblock Norm-based capacity control in neural networks, 2015{\natexlab{b}}.

\bibitem[Pang et~al.(2022)Pang, Lin, Yang, Zhu, and Yan]{Pang2022RobustnessAA}
Tianyu Pang, Min Lin, Xiao Yang, Junyi Zhu, and Shuicheng Yan.
\newblock Robustness and accuracy could be reconcilable by (proper) definition.
\newblock In \emph{International Conference on Machine Learning}, 2022.

\bibitem[Raghunathan et~al.(2018)Raghunathan, Steinhardt, and Liang]{raghunathan2018certified}
Aditi Raghunathan, Jacob Steinhardt, and Percy Liang.
\newblock Certified defenses against adversarial examples.
\newblock \emph{arXiv preprint arXiv:1801.09344}, 2018.

\bibitem[Rice et~al.(2020)Rice, Wong, and Kolter]{rice2020overfitting}
Leslie Rice, Eric Wong, and J.~Zico Kolter.
\newblock Overfitting in adversarially robust deep learning, 2020.

\bibitem[Ross and Doshi-Velez(2017)]{ross2017improving}
Andrew~Slavin Ross and Finale Doshi-Velez.
\newblock Improving the adversarial robustness and interpretability of deep neural networks by regularizing their input gradients, 2017.

\bibitem[Szegedy et~al.(2013)Szegedy, Zaremba, Sutskever, Bruna, Erhan, Goodfellow, and Fergus]{szegedy2013intriguing}
Christian Szegedy, Wojciech Zaremba, Ilya Sutskever, Joan Bruna, Dumitru Erhan, Ian Goodfellow, and Rob Fergus.
\newblock Intriguing properties of neural networks.
\newblock \emph{arXiv preprint arXiv:1312.6199}, 2013.

\bibitem[Tsipras et~al.(2019)Tsipras, Santurkar, Engstrom, Turner, and Madry]{tsipras2019robustness}
Dimitris Tsipras, Shibani Santurkar, Logan Engstrom, Alexander Turner, and Aleksander Madry.
\newblock Robustness may be at odds with accuracy, 2019.

\bibitem[Wang et~al.(2022)Wang, Zhang, Xu, and Ruan]{wang2022deep}
Fu Wang, Chi Zhang, Peipei Xu, and Wenjie Ruan.
\newblock Deep learning and its adversarial robustness: A brief introduction.
\newblock In \emph{HANDBOOK ON COMPUTER LEARNING AND INTELLIGENCE: Volume 2: Deep Learning, Intelligent Control and Evolutionary Computation}, pages 547--584. 2022.

\bibitem[Wang et~al.(2023{\natexlab{a}})Wang, Fu, Zhang, and Ruan]{wang2023self}
Fu Wang, Zeyu Fu, Yanghao Zhang, and Wenjie Ruan.
\newblock Self-adaptive adversarial training for robust medical segmentation.
\newblock In \emph{International Conference on Medical Image Computing and Computer-Assisted Intervention (MICCAI’23)}, pages 725--735. Springer, 2023{\natexlab{a}}.

\bibitem[Wang et~al.(2020)Wang, Zou, Yi, Bailey, Ma, and Gu]{mart}
Yisen Wang, Difan Zou, Jinfeng Yi, James Bailey, Xingjun Ma, and Quanquan Gu.
\newblock Improving adversarial robustness requires revisiting misclassified examples.
\newblock In \emph{International Conference on Learning Representations}, 2020.

\bibitem[Wang and Ruan(2022)]{wang2022understanding}
Zheng Wang and Wenjie Ruan.
\newblock Understanding adversarial robustness of vision transformers via cauchy problem.
\newblock In \emph{Joint European Conference on Machine Learning and Knowledge Discovery in Databases (ECML/PKDD'22)}, 2022.

\bibitem[Wang et~al.(2023{\natexlab{b}})Wang, Pang, Du, Lin, Liu, and Yan]{wang2023better}
Zekai Wang, Tianyu Pang, Chao Du, Min Lin, Weiwei Liu, and Shuicheng Yan.
\newblock Better diffusion models further improve adversarial training.
\newblock \emph{arXiv preprint arXiv:2302.04638}, 2023{\natexlab{b}}.

\bibitem[Wu et~al.(2020)Wu, Xia, and Wang]{wu2020adversarial}
Dongxian Wu, Shu-Tao Xia, and Yisen Wang.
\newblock Adversarial weight perturbation helps robust generalization.
\newblock \emph{Advances in Neural Information Processing Systems}, 33:\penalty0 2958--2969, 2020.

\bibitem[Wu et~al.(2023)Wu, Yunas, Rowlands, Ruan, and Wahlstr{\"o}m]{wu2023advdriving}
Han Wu, Syed Yunas, Sareh Rowlands, Wenjie Ruan, and Johan Wahlstr{\"o}m.
\newblock Adversarial driving: Attacking end-to-end autonomous driving.
\newblock In \emph{2023 IEEE Intelligent Vehicles Symposium (IV)}, pages 1--7. IEEE, 2023.

\bibitem[Yang et~al.(2020)Yang, Rashtchian, Zhang, Salakhutdinov, and Chaudhuri]{yang2020closer}
Yao-Yuan Yang, Cyrus Rashtchian, Hongyang Zhang, Ruslan Salakhutdinov, and Kamalika Chaudhuri.
\newblock A closer look at accuracy vs. robustness, 2020.

\bibitem[Yin et~al.(2022)Yin, Ruan, and Fieldsend]{yin2022dimba}
Xiangyu Yin, Wenjie Ruan, and Jonathan Fieldsend.
\newblock Dimba: discretely masked black-box attack in single object tracking.
\newblock \emph{Machine Learning}, pages 1--19, 2022.

\bibitem[Yin et~al.(2023)Yin, Wu, Liu, Fang, Zhao, Huang, and Ruan]{yin2023rerogcrl}
Xiangyu Yin, Sihao Wu, Jiaxu Liu, Meng Fang, Xingyu Zhao, Xiaowei Huang, and Wenjie Ruan.
\newblock Rerogcrl: Representation-based robustness in goal-conditioned reinforcement learning.
\newblock \emph{arXiv preprint arXiv:2312.07392}, 2023.

\bibitem[Yu et~al.(2021)Yu, Yang, Dobriban, Steinhardt, and Ma]{yu2021understanding}
Yaodong Yu, Zitong Yang, Edgar Dobriban, Jacob Steinhardt, and Yi Ma.
\newblock Understanding generalization in adversarial training via the bias-variance decomposition, 2021.

\bibitem[Zhang et~al.(2023)Zhang, Ruan, and Xu]{zhang2023reachability}
Chi Zhang, Wenjie Ruan, and Peipei Xu.
\newblock Reachability analysis of neural network control systems.
\newblock In \emph{Proceedings of the AAAI Conference on Artificial Intelligence (AAAI'23)}, 2023.

\bibitem[Zhang et~al.(2019{\natexlab{a}})Zhang, Yu, Jiao, Xing, Ghaoui, and Jordan]{Zhang2019TheoreticallyPT}
Hongyang Zhang, Yaodong Yu, Jiantao Jiao, Eric~P. Xing, Laurent~El Ghaoui, and Michael~I. Jordan.
\newblock Theoretically principled trade-off between robustness and accuracy.
\newblock \emph{ArXiv}, abs/1901.08573, 2019{\natexlab{a}}.

\bibitem[Zhang et~al.(2019{\natexlab{b}})Zhang, Yu, Jiao, Xing, Ghaoui, and Jordan]{zhang2019theoretically}
Hongyang Zhang, Yaodong Yu, Jiantao Jiao, Eric~P. Xing, Laurent~El Ghaoui, and Michael~I. Jordan.
\newblock Theoretically principled trade-off between robustness and accuracy, 2019{\natexlab{b}}.

\end{thebibliography}
}

\clearpage
\setcounter{page}{1}
\maketitlesupplementary
In this appendix, we provide detailed descriptions of the methodologies and experimental procedures used in our study, which encompasses:

\begin{itemize}
\item Theoretical proofs in detail.
\item Additional experimental settings and results.
\item Limitations and future directions.
\end{itemize}

The aim is to ensure the transparency and reproducibility of our research, while providing sufficient information for readers interested in the technical intricacies of our work.
\section{Proofs of Lemma.~\ref{frn} and Thm.~\ref{frn-ce}}
According to Lemma.~\ref{fisher-rao}, we can denote the Fisher-Rao norm of $\mathcal{W}$ w.r.t $\mathcal{L}_{ce}$ as:
\begin{align}
\small
\begin{split}
\left\|\mathcal{W}\right\|_{FR\circ \mathcal{L}_{ce}}^2 &= L^2\mathbb{E}_{(\mathbf{x}, y)}\left[\left(\langle \sigma(f_{\mathcal{W}}^{L}(\mathbf{x})), f^{L}_{\mathcal{W}}(\mathbf{x})\rangle-f_{\mathcal{W}}^{L}(\mathbf{x})_y\right)^2\right]\\
\end{split}
\end{align}
Then we can conclude that 
\begin{align}
\begin{split}
\gamma_{ce}&=\mathbb{E}_{(\mathbf{x},y)}\left[\left|\langle \sigma(f^{L}_{\mathcal{W}}(\mathbf{x})),f^{L}_{\mathcal{W}}(\mathbf{x})\rangle - f^L_{\mathcal{W}}(\mathbf{x})_y\right|\right]\\
&=\mathbb{E}_{(\mathbf{x},y)}\left[\left|\sum_{k=1}^{K}\frac{(f^L_{\mathcal{W}}(\mathbf{x})_k-f^L_{\mathcal{W}}(\mathbf{x})_y)e^{f^L_{\mathcal{W}}(\mathbf{x})_k}}{\sum_{k=1}^{K}e^{f^L_{\mathcal{W}}(\mathbf{x})_k}}\right|\right]\\
&=\mathbb{E}_{(\mathbf{x},y)}\left[\left|\sum_{k\neq y}\frac{(f^L_{\mathcal{W}}(\mathbf{x})_k-f^L_{\mathcal{W}}(\mathbf{x})_y) e^{f^L_{\mathcal{W}}(\mathbf{x})_k}}{\sum_{k=1}^{K}e^{f^L_{\mathcal{W}}(\mathbf{x})_k}}\right|\right]\\
&\leq\mathbb{E}_{(\mathbf{x},y)}\left[\frac{\sum_{k\neq y}\left|(f^L_{\mathcal{W}}(\mathbf{x})_k-f^L_{\mathcal{W}}(\mathbf{x})_y)\right| e^{f^L_{\mathcal{W}}(\mathbf{x})_k}}{\sum_{k\neq y}e^{f^L_{\mathcal{W}}(\mathbf{x})_k}}\right]\\
&\leq\mathbb{E}_{(\mathbf{x}, y)}\left[\max_{k\neq y}\left|f^L_{\mathcal{W}}(\mathbf{x})_k - f^L_{\mathcal{W}}(\mathbf{x})_y\right|\right]\\
\end{split}
\end{align}
Then the standard Rademacher complexity w.r.t the CE loss $\mathcal{R}_{N_{tr}}(\mathcal{L}_{ce}\circ \mathcal{F}_{\hat{\gamma}_{ce}})$ can be denoted as:
\begin{align}
\begin{split}
&\mathbb{E}_{\xi}\sup_{f_{\mathcal{W}}^L\in\mathcal{F}_{\hat{\gamma}_{ce}}}\frac{1}{N}\sum_{i=1}^{N}\xi_{i}\mathcal{L}_{ce}\left(f_{\mathcal{W}}^L(\mathbf{x}_i), y_i\right)\\
&=\mathbb{E}_{\xi}\sup_{f^L_{\mathcal{W}}\in\mathcal{F}_{\hat{\gamma}_{ce}}}\frac{1}{N}\sum_{i=1}^{N}\xi_{i}\ln\left(\frac{e^{f^L_{\mathcal{W}}(\mathbf{x}_i)_y}}{\sum_{k=1}^{K}e^{f^L_{\mathcal{W}}(\mathbf{x}_i)_k}}\right)\\
&=\mathbb{E}_{\xi}\sup_{f_{\mathcal{W}}^L\in\mathcal{F}_{\hat{\gamma}_{ce}}}\frac{1}{N}\sum_{i=1}^{N}\xi_{i}\left(\ln\left(\sum_{k=1}^Ke^{f^L_{\mathcal{W}}(\mathbf{x}_i)_k}\right)-f^L_{\mathcal{W}}(\mathbf{x}_i)_y\right)\\
&=\mathbb{E}_{\xi}\sup_{f_{\mathcal{W}}^L\in\mathcal{F}_{\hat{\gamma}_{ce}}}\frac{1}{N}\sum_{i=1}^{N}\xi_{i}\left(\ln\left(\sum_{k=1}^K\left(\frac{1}{e}\right)^{f^L_{\mathcal{W}}(\mathbf{x}_i)_y-f^L_{\mathcal{W}}(\mathbf{x}_i)_k}\right)\right)\\
\end{split}
\end{align}
Specifically, for $N_{tr}^{M}$ misclassified samples, the upper bound for $\mathcal{R}_{N_{tr}^M}\left(\mathcal{L}_{ce}\circ\mathcal{F}_{\hat{\gamma}_{ce}}\right)$ can be further derived as:
\begin{align}
\label{misclassified}
\begin{split}
&\mathcal{R}_{N_{tr}^M}\left(\mathcal{L}_{ce}\circ \mathcal{F}_{\hat{\gamma}_{ce}}\right)\\
&\leq\mathbb{E}_{\xi}\sup_{f_{\mathcal{W}}^L\in\mathcal{F}_{\hat{\gamma}_{ce}}}\frac{1}{N_{tr}^{M}}\sum_{i=1}^{N_{tr}^{M}}\xi_{i}\left(\ln\left(\sum_{k=1}^K\left(\frac{1}{e}\right)^{f_{\mathcal{W}}^L(\mathbf{x}_i)_y - f_{\mathcal{W}}^L(\mathbf{x}_i)_k}\right)\right)\\
&\leq\mathbb{E}_{\xi}\frac{1}{N_{tr}^{M}}\sum_{i=1}^{N_{tr}^{M}}\xi_{i}\left(\ln\left(\sum_{k=1}^K\left(\frac{1}{e}\right)^{-\left|f_{\mathcal{W}}^L(\mathbf{x}_i)_y - f_{\mathcal{W}}^L(\mathbf{x}_i)_k\right|}\right)\right)\\
&\leq\left(\ln K+\hat{\gamma}_{ce}\right)\sqrt{\frac{1}{{N_{tr}^{M}}^2}(\mathbb{E}_{\xi}\sum_{i=1}^{N_{tr}^M}\xi_i)^2}\\
&\leq\left(\ln K+\hat{\gamma}_{ce}\right)\sqrt{\frac{1}{N_{tr}^{M}}(\mathbb{E}_{\xi}\sum_{i=1}^{N_{tr}^M}\xi_i)}\\
&\stackrel{(i)}{\lesssim}\frac{\ln K+\hat{\gamma}_{ce}}{\sqrt{N_{tr}^M}}
\end{split}
\end{align}
Notice that it is very likely that $0<\left|\mathbb{E}_{\xi}\sum_{i=1}^{N_{tr}^M}\xi_i\right|<N_{tr}^{M}$, we then assume $\left|\mathbb{E}_{\xi}\sum_{i=1}^{N_{tr}^M}\xi_i\right|\approx\sqrt{N_{tr}^M}$. Due to the fact that we are deriving the upper bound, (i) holds. Meanwhile, for $N_{tr}^C$ correctly classified samples, the following inequalities holds:
\begin{align}
\begin{split}
&\mathcal{R}_{N_{tr}^C}\left(\mathcal{L}_{ce}\circ \mathcal{F}_{\hat{\gamma}_{ce}}\right)\\
&\geq\mathbb{E}_{\xi}\frac{1}{N_{tr}^{C}}\sum_{i=1}^{N_{tr}^{C}}\xi_{i}\left(\ln\left(\sum_{k=1}^K\left(\frac{1}{e}\right)^{\hat{\gamma}_{ce}}\right)\right)\\
&\geq\frac{(\ln K - \hat{\gamma}_{ce})}{N_{tr}^C}\mathbb{E}_{\xi}\sum_{i=1}^{N_{tr}^C}\xi_i\\
\end{split}
\end{align}
Similar to Eq.~\ref{misclassified}, we assume $\left|\mathbb{E}_{\xi}\sum_{i=1}^{N_{tr}^C}\xi_i\right|\approx\sqrt{N_{tr}^C}$. Meanwhile, due to variations in network architecture, training data, and training algorithm, etc, the value of  $\hat{\gamma}_{ce}$ tends to vary a lot, therefore, we need to make $\mathbb{E}_{\xi}\sum_{i=1}^{N_{tr}^{C}}\xi_i$ positive to reach the lower bound, which can be denoted as $\mathcal{R}_{N_{tr}^C}\left(\mathcal{L}_{ce}\circ \mathcal{F}_{\hat{\gamma}_{ce}}\right)\gtrsim\frac{\ln K - \hat{\gamma}_{ce}}{\sqrt{N_{tr}^C}}$. Consequently, we can deduce that:
\begin{align}
\begin{split}
\mathcal{R}_{N_{tr}}\left(\mathcal{L}_{ce}\circ\mathcal{F}_{\hat{\gamma}_{ce}}\right)&\leq\frac{(\sqrt{N_{tr}^{M}}+\sqrt{N_{tr}^C})\ln K + \hat{\gamma}_{ce}\sqrt{N_{tr}^M}}{N_{tr}}\\
\mathcal{R}_{N_{tr}}\left(\mathcal{L}_{ce}\circ\mathcal{F}_{\hat{\gamma}_{ce}}\right)&\geq\frac{(\sqrt{N_{tr}^{M}}+\sqrt{N_{tr}^C})\ln K - \hat{\gamma}_{ce}\sqrt{N_{tr}^C}}{N_{tr}}
\end{split}
\end{align}
Then if we define $\Gamma_{ce}=\frac{\hat{\gamma}_{ce}^{N_{tr}^C}-\hat{\gamma}_{ce}^{N_{tr}^M}}{\hat{\gamma}_{ce}^{N_{tr}^M}}$, $\mathcal{C}_{C}=\frac{N_{tr}}{N_{tr}^C}$, $\mathcal{C}_{M}=\frac{N_{tr}}{N_{tr}^{M}}$, $\mathcal{C}_{MC}=\frac{\sqrt{N_{tr}^{C}}+\sqrt{N_{tr}^M}}{N_{tr}}$, we can easily provide the following bounds:
\begin{align}
\begin{split}
\mathcal{R}_{N_{tr}}\left(\mathcal{L}_{ce}\circ \mathcal{F}_{\hat{\gamma}_{ce}}\right)&\leq\mathcal{C}_{MC}\ln K + \frac{\hat{\gamma}_{ce}^{N_{tr}^M}\mathcal{C}_{M}^{-0.5}(\mathcal{C}_{C}^{-1}\Gamma_{ce}+1)}{N_{tr}^{0.5}}\\
\mathcal{R}_{N_{tr}}\left(\mathcal{L}_{ce}\circ \mathcal{F}_{\hat{\gamma}_{ce}}\right)&\geq\mathcal{C}_{MC}\ln K - \frac{\hat{\gamma}_{ce}^{N_{tr}^M}\mathcal{C}_{C}^{-0.5}(\mathcal{C}_{C}^{-1}\Gamma_{ce}+1)}{N_{tr}^{0.5}}\\
\end{split}
\end{align}
\section{Additional Experimental Settings and Results}
\textbf{Experimental Settings.} The qualitative Results illustrated in Fig.~\ref{figure:complexity} and \ref{figure:lambda} are derived from training models on a dataset comprising 4000 samples from MNIST. For the evaluations presented in Fig.~\ref{figure:width-wise}, we utilized 4,000 training instances from MNIST and 10,000 from the CIFAR10. Furthermore, for all setups in LOAT-boosted training, we consistently set the parameters $\tau$ to 1 and $\gamma$ to 0.05, following the specifications of Alg.~\ref{alg}.

\textbf{Additional Results.}
In addition to the LOAT-boosted S2O results on ResNet18, as detailed in Tab.~\ref{tab:s2o}, we extended our experiments to WideResNet-34-10. 
\begin{table}[htbp]
\centering
\scriptsize 
\setlength{\tabcolsep}{2pt} 
\renewcommand{\arraystretch}{1.0}
\caption{Evaluation of LOAT-boosted S2O on WideResNet-34-10 (\%).}
\begin{tabular}{lllll}
\toprule
Defense &  Clean$_{te}$  & PGD$^{7}$ & PGD$^{20}$ & PGD$^{40}$ \\
\midrule
  PGD-AT & 85.57 & 53.20 & 47.84 & 46.78 \\
  PGD-AT+SLORE &  \textbf{{86.97}} &   \textit{{53.51}}     &  \textit{{48.20}} &  \textit{{47.23}}  \\
  PGD-AT+LORE &    \textit{{86.69}}   &   \textbf{{53.90}}  & \textbf{{48.77}} & \textbf{{47.82}} \\
\bottomrule
\end{tabular}
\label{s2o_addition}
\end{table}
\begin{table}[htbp]
\centering
\scriptsize 
\setlength{\tabcolsep}{2pt} 
\renewcommand{\arraystretch}{1.0}
\caption{Comparison of SLORE-boosted PGD-AT with ALP-boosted one on ResNet-18 (\%).}
\begin{tabular}{lllll}
\toprule
Defense &  Clean$_{te}$  & FGSM & PGD$^{7}$ & PGD$^{20}$ \\
\midrule
  PGD-AT (ALP) & 81.26 & 60.92 & 51.38 & 45.90 \\
  PGD-AT (SLORE) &  \textbf{{83.41}} &   \textbf{{61.80}}     &  \textbf{{51.46}} &  \textbf{{46.50}} \\
\bottomrule
\end{tabular}
\label{alp_comparison}
\end{table}
The comprehensive outcomes of these additional experiments are presented in Tab.~\ref{s2o_addition}.  
\begin{table}
\scriptsize
\setlength{\tabcolsep}{2pt}
\centering
\caption{Classification Accuracy of models trained by 1$\times 10^6$ EDM-generated images-augmented SVHN (\%).}
\begin{tabular}{c|c|c|c|c|c|c}
\toprule
Models & Defense & Clean$_{tr}$ & Clean$_{te}$ &PGD$^{20}$ & PGD$^{40}$  & AA  \\
\midrule
\multirow{6}{*}{\centering\makecell{PreResNet18 \\ (Swish)}}   & TRADES & 96.59 & 96.57 & 56.74 & 53.84 & 36.21\\
                              & TRADES+SLORE
                              & 96.22
                              & 96.14 & \textbf{61.20} & \textbf{57.85} & \textbf{38.58} \\
                              & TRADES+LORE
                              & \textbf{97.79}
                              & \textbf{97.39} & 55.65 & 47.33 & 9.64\\
                 \cmidrule(lr){2-7}
                             & TRADES(S)
                             &
                             96.06
                             & 96.20  & 69.91 & 66.64& \textbf{41.38}\\
                             & TRADES(S)+SLORE 
                             &
                            \textbf{97.12} & \textbf{97.01}
                             & \textbf{77.73} & \textbf{73.38} & 13.48\\
& TRADES(S)+LORE 
                             &
                             96.46
                             & 96.44 & 65.25& 61.82& 37.65\\
\midrule
\multirow{6}{*}{\centering\makecell{WRN-28-10 \\ (Swish)}}  &TRADES & 97.45 & \textbf{97.69}  & 59.57 & 54.75 & 15.17\\ 
                             & TRADES+SLORE 
                             & 95.46 & 95.67 & \textbf{68.35} & \textbf{67.01} & \textbf{49.25}\\
                             & TRADES+LORE 
                             & \textbf{97.61} & 97.66  & 43.44 & 37.51 & 10.79 \\
                 \cmidrule(lr){2-7}
                            & TRADES(S) 
                            & \textbf{97.43}
                            & \textbf{97.75}  & \textbf{74.98} & \textbf{69.10} & 15.77\\
                            & TRADES(S)+SLORE
                            & 97.11 & 97.31 & 70.17 & 65.19 & 11.27
                            \\
                            & TRADES(S)+LORE
                            & 97.22 & 97.36 & 57.71 & 50.58 & \textbf{16.36}
                            \\
                 \cmidrule(lr){2-7} 
                            & MART 
                            & \textbf{97.23}
                            & \textbf{97.23}  & 52.34 & 49.89 & 22.89\\
                             & MART+SLORE 
                             & 88.13
                             & 93.97  & 64.22 & 63.82 & \textbf{53.26}\\
& MART+LORE
                            & 94.70
                            & 96.30 & \textbf{72.16} & \textbf{70.76} & 9.03\\
                            
\bottomrule
\end{tabular}
\label{tab_edm_svhn}
\end{table}
Our findings indicate that SLORE-enhanced PGD-AT can enhance the test clean accuracy of WideResNet-34-10 by 1.40\%. Furthermore, LORE-enhanced PGD-AT contributes to an increase in the model's robustness against PGD$^7$ attacks by 0.70\%, PGD$^{20}$ attacks by 0.93\%, and PGD$^{40}$ attacks by 1.04\%.

Furthermore, we compare our algorithm with widely used Adversarial Logit Pairing (ALP). Detailed results are shown in Tab.~\ref{alp_comparison}. 

Additionally, we conducted evaluations of the LOAT-boosted DM-AT on augmented datasets from Cifar100, Tiny-ImageNet and SVHN. The extensive results in Tab.~\ref{tab_edm_svhn}, \ref{tab_edm_cifar100} and \ref{tab_edm_tinyimagenet} reinforce the adaptability and efficacy of our approach.
\begin{table}
\scriptsize
\setlength{\tabcolsep}{2pt}
\centering
\caption{Classification Accuracy of models trained by 1$\times 10^6$ EDM-generated images-augmented Cifar100 (\%).}
\begin{tabular}{c|c|c|c|c|c|c}
\toprule
Models & Defense & Clean$_{tr}$ & Clean$_{te}$ &PGD$^{20}$ & PGD$^{40}$  & AA  \\
\midrule
\multirow{6}{*}{\centering\makecell{PreResNet18 \\ (Swish)}}   & TRADES & 85.27 & 62.33 & 34.62 & \textbf{34.59} & \textbf{29.66} \\
                              & TRADES+SLORE
                              & 84.95
                              & 61.63 & 34.41 & 34.47 & 29.40 \\
                              & TRADES+LORE
                              & \textbf{86.35}
                              & \textbf{62.79} & \textbf{34.63} & 34.55 & 29.21\\
                 \cmidrule(lr){2-7}
                             & TRADES(S)
                             &
                             81.84
                             & 60.15  & 33.82& 33.76& 28.76\\
                             & TRADES(S)+SLORE 
                             &
                            \textbf{81.85} & \textbf{60.22}
                             & 33.66 & 33.73 & \textbf{28.88}\\
& TRADES(S)+LORE 
                             &
                             81.81
                             & 59.86 & \textbf{34.02}& \textbf{34.00}& 28.70\\
                 \cmidrule(lr){2-7}
                             & MART 
                             &
                             77.78
                             & 59.91   & 34.86 & 34.83 & 29.18\\
                             & MART+SLORE 
                             &
                             77.08
                             & 59.86  &34.72& 34.70 & \textbf{29.60}\\
& MART+LORE 
                             &
                             \textbf{78.57}&\textbf{60.18}
                             &\textbf{35.20}& \textbf{35.22} & 29.47\\
                             
\midrule
\multirow{6}{*}{\centering\makecell{WRN-28-10 \\ (Swish)}}  &TRADES & 89.39 & 65.41   & 37.88 & 37.93 & 33.02\\ 
                             & TRADES+SLORE 
                             & 88.95&64.95 & 38.07 & 38.04 & \textbf{33.29} \\
                             & TRADES+LORE 
                             & \textbf{90.17} & \textbf{66.62}  & \textbf{38.10} & \textbf{37.96} & 32.61\\
                 \cmidrule(lr){2-7}
                            & TRADES(S) 
                            & \textbf{85.22}
                            & 63.12  & 36.62 & 36.57 & 31.60 \\
                            & TRADES(S)+SLORE
                            &84.78&\textbf{63.18} & 36.50 & 36.54& \textbf{31.97}                            \\
                            & TRADES(S)+LORE
                            &84.82&62.91 & \textbf{36.78} & \textbf{36.65}&  31.56
                            \\
                 \cmidrule(lr){2-7} 
                            & MART 
                            & 81.82
                            & 65.05  & 38.56 & 38.51 & 33.91 \\
                             & MART+SLORE 
                             & 80.35
                             & 64.46  & 38.88 & 38.81 & \textbf{34.26}\\
& MART+LORE
                            & \textbf{83.21}
                            & \textbf{65.11}  & \textbf{39.06} & \textbf{39.00} & 33.76\\
                            
\bottomrule
\end{tabular}
\label{tab_edm_cifar100}
\end{table}
\begin{table}
\small
\centering
\setlength\tabcolsep{0.5pt}
\caption{Classification Accuracy of PreResNet18 (Swish) trained by  1$\times$$10^{6}$ EDM-generated images-augmented Tiny-ImageNet (\%). V, S, and L represent vanilla, SLORE-boosted, and LORE-boosted adversarial training algorithms respectively. C$_{tr}$, C$_{te}$, P$_{40}$, A$_{ce}$, and A$_{t}$ indicate Clean$_{tr}$, Clean$_{te}$, PGD$^{40}$, APGD$_{ce}$, and APGD$_{t}$ individually. }
\label{tiny-imagenet}
\begin{tabular}{|c|c|c|c|c|c|c|c|c|c|c|}
\hline
\multirow{2}{*}{} & \multicolumn{5}{c|}{TRADES} & \multicolumn{5}{c|}{MART} \\ \cline{2-11} 
                       & C$_{tr}$ & C$_{te}$ & P$_{40}$ & A$_{ce}$ & A$_T$ & C$_{tr}$ & C$_{te}$ & P$_{40}$ & A$_{ce}$ &A$_T$ \\ 
\cline{1-11}
V              &    64.91     &   58.66    &  28.52    &    28.31     &   \textbf{21.96}    &  52.52    &  50.62       &   28.14    &  28.02  & 21.51 \\
\cline{1-1}
S              &    64.76     &   58.32    & 28.33    &    28.32     &   21.59    &  \textbf{55.17}    &    \textbf{51.34}     &   \textbf{29.02}    &  \textbf{28.75}  & \textbf{22.54} \\ 
\cline{1-1}
L             &  \textbf{65.23}      &  \textbf{59.08}    &   \textbf{28.78}   &    \textbf{28.47}     &   21.68    &  53.04    &  50.79       &  28.61     &  27.74 & 21.26  \\ \hline
\end{tabular}
\label{tab_edm_tinyimagenet}
\end{table}
To more comprehensively demonstrate the efficacy of our approach, 
\begin{table}[htbp]
\centering
\scriptsize 
\setlength{\tabcolsep}{2pt} 
\renewcommand{\arraystretch}{1.0}
\caption{Classification Accuracy of PreResNet18 (Swish) trained by  1$\times$$10^{6}$ EDM-generated images-augmented Cifar10  after 400 epochs (\%).}
\begin{tabular}{llllll}
\toprule
Defense &  Clean$_{te}$  & PGD$^{10}$ & PGD$^{20}$ & PGD$^{40}$ & AA\\
\midrule
  TRADES & 88.45 & 62.62 & 61.88 & 61.76 & \textbf{58.21}\\
  TRADES+SLORE &  86.12 &   62.28    &  61.73 &  61.56 &  57.76 \\
  TRADES+LORE &  \textbf{90.26}   &   \textbf{62.81}  & \textbf{62.01} & \textbf{61.85} & 56.51\\

\cmidrule(lr){1-6}

  MART  & 88.35 & 61.19 & 60.18 & 59.93 & 53.92\\
  MART+SLORE &  88.15 &   61.99   & 60.94 & 60.78 & 54.74\\
  MART+LORE &  \textbf{89.08}   &   \textbf{63.81}  & \textbf{62.83} & \textbf{62.57} & \textbf{55.28} \\
\bottomrule
\end{tabular}
\label{epoch_400}
\end{table}
we trained LOAT-boosted TRADES and MART on PreResNet18 (Swish) for an extended duration of 400 epochs. The evaluations, as presented in Tab.~\ref{epoch_400}, reveal a more pronounced improvement in performance compared to training for approximately 100 epochs. This enhancement is particularly notable in the case of MART.

\section{Limitations and Future Works}

\itemize

\item \textbf{Mitigating Theoretical Gap Between MLPs and Other Network Structures.}
We intend to implement knowledge distillation using a ReLU-activated MLP, aligning its architecture to mirror the layer count of a ResNet residual block or a Transformer stack, and matching the hidden units in each layer to the dimensionality of latent features derived from these structures. This framework will be employed in LOAT-boosted AT algorithms, which can help us investigate the theoretical gap.

\item \textbf{Hyperparameters in LOAT.} 
The selection of hyperparameters $\mathcal{E}_1$, $\mathcal{E}_2$, $\tau$, and $\gamma$ is crucial, as their configuration can significantly impact the final results.

\item \textbf{Future Directions.}
The rising prominence of Transformers and Large Language Models (LLMs) presents new challenges and opportunities. Exploring the application of adversarial training frameworks to these models is both an intriguing and vital avenue for future research. 

\label{sec:appendix}

\end{document}